\documentclass[sn-mathphys-num]{sn-jnl}

\usepackage{graphicx}%
\usepackage{mathrsfs} 
\usepackage{multirow}%
\usepackage{amsmath,amssymb,amsfonts}%
\usepackage{amsthm}%
\usepackage{mathrsfs}%
\usepackage[title]{appendix}%
\usepackage{xcolor}%
\usepackage{textcomp}%
\usepackage{manyfoot}%
\usepackage{booktabs}%
\usepackage{algorithm}%
\usepackage{algorithmicx}%
\usepackage{algpseudocode}%
\usepackage{listings}%

\theoremstyle{thmstyleone}%
\theoremstyle{thmstyletwo}%

\theoremstyle{thmstylethree}%

\begin{document}
\title[Article Title]{Adaptive parameters identification for nonlinear dynamics  using deep permutation invariant networks}


\author*[1,2]{\fnm{Mouad} \sur{ELAARABI}}\email{mouad.elaarabi@ec-nantes.fr, mouad.el-aarabi@irt-jules-verne.fr}

\author[1]{\fnm{Domenico} \sur{BORZACCHIELLO}}\email{domenico.borzacchiello@ec-nantes.fr}

\author[2]{\fnm{Philippe} \sur{LE BOT}}\email{philippe.le-bot@irt-jules-verne.fr}

\author[2]{\fnm{Yves} \sur{LE GUENNEC}}\email{yves.le-guennec@irt-jules-verne.fr}

\author[1]{\fnm{Sebastien} \sur{COMAS-CARDONA}}\email{sebastien.comas@ec-nantes.fr}

\affil[1]{\orgdiv{Nantes Université}, \orgname{Ecole Centrale Nantes, CNRS, GeM, UMR 6183}, \orgaddress{\city{Nantes}, \postcode{44321}, \country{France}}}

\affil[2]{\orgdiv{Nantes Université}, \orgname{IRT Jules Verne}, \orgaddress{\city{Bouguenais}, \postcode{44340}, \country{France}}}



\abstract{The promising outcomes of dynamical system identification techniques, such as SINDy~\cite{SINDy}, highlight their advantages in providing qualitative interpretability and extrapolation compared to non-interpretable deep neural networks~\cite{rudin2019stop}. These techniques suffer from parameter updating in real-time use cases, especially when the system parameters are likely to change during or between processes. Recently, the OASIS~\cite{OASIS} framework introduced a data-driven technique to address the limitations of real-time dynamical system parameters updating, yielding interesting results. Nevertheless, we show in this work that superior performance can be achieved using more advanced model architectures.\\
We present an innovative encoding approach, based mainly on the use of Set Encoding methods of sequence data, which give accurate adaptive model identification for complex dynamic systems, with variable input time series length. Two Set Encoding methods are used: the first is Deep Set~\cite{DEEPSETS}, and the second is Set Transformer~\cite{SETTRANS}. Comparing Set Transformer to OASIS framework on Lotka-Volterra for real-time local dynamical system identification and time series forecasting, we find that the Set Transformer architecture is well adapted to learning relationships within data sets. We then compare the two Set Encoding methods based on the Lorenz system for online global dynamical system identification. \textcolor{black}{Finally, we trained a Deep Set model to perform identification and characterization of abnormalities for 1D heat-transfer problem.}}





\maketitle

\section{Introduction}
Exploring dynamic systems through differential equations is intriguing, providing valuable insights into variable interactions and system behavior. These insights can be useful for many tasks, such as monitoring and controlling industrial processes. However, this could be a challenging undertaking due to their inherent complexity (such as variability of parameters and representativeness of the model). The direct construction of these equations from theoretical principles is a significant challenge and may be sensitive to changes in the parameters or modeling hypotheses, leading researchers to rely on alternative methods such as analyzing empirical data through simulations or experiments. This process involves generating datasets and conducting meticulous analyses using appropriate methods, offering valuable insights about the intrinsic interactions between variables and enhancing the understanding of physical system dynamics. \\\\
While machine learning methods offer a potential solution, they suffer from issues such as a lack of interpretability, as the trained model is a black box and generally requires extensive training data for high accuracy, which is not always readily available. These methods may also struggle to generalize for unseen data~\cite{rudin2019stop}. On the other hand, model discovery methods~\cite{SID_SUR, lejeune2023data} aims to identify the true governing equation through empirical data, providing insights into the interactions between the data, which could be applied to a wide range of engineering tasks.\\

\begin{table}[h]
\caption{Abbreviation table }\label{tab:abb_table}
\begin{tabular}{@{}ll@{}}
\toprule
\textbf{Abbreviation}  & \textbf{Definition} \\
\midrule
CAE & Computer-Aided Engineering\\\hline
DMD & Dynamic Mode Decomposition \\\hline
DMDc & Dynamic Mode Decomposition with control \\\hline
DNN & Deep Neural Network \\\hline
EQL & EQuation Learning \\\hline
ISAB & Induced Self-Attention Block \\\hline
LASSO & Least Absolute Shrinkage and Selection Operator \\\hline
MAB & Multi-head Attention Block \\\hline
MAPE & Mean Absolute Percentage Error \\\hline
MLP & MultiLayer Perceptron \\\hline
MPC & Model Predictive Control \\\hline
NSODE & Neural Symbolic Ordinary Differential Equation \\\hline
OASIS & Operable Adaptive Sparse Identification of Systems \\\hline
OASIS-P & Online Adaptive Sparse Identification of Systems for Fault Prognosis \\\hline
ODE & Ordinary Differential Equation \\\hline
PINN & Physics Informed Neural Network \\\hline
PINN-SR & Physics Informed Neural Network with Sparse Regression \\\hline
PDE & Partial Differential Equation \\\hline
PMA & Pooling by Multihead Attention \\\hline
PySR & Python Symbolic Regression \\\hline
SAB & Self-Attention Block \\\hline
SENDI & Set Encoding for Nonlinear Dynamics Identification \\\hline
SGA-PDE & Symbolic Genetic Algorithm for discovering open-form Partial Differential Equations \\\hline
SINDYc & Sparse Identification of Nonlinear Dynamics with Control \\\hline
SINDy & Sparse Identification of Nonlinear Dynamics \\\hline
SR3 & Sparse Relaxed Regularized Regression \\\hline
rFF & row-wise FeedForward \\\hline
sMAPE & symmetric Mean Absolute Percentage Error \\
\botrule
\end{tabular}
\end{table}

\subsection{Background and motivations}

In this work, we focus on dynamic system identification methods~\cite{NDY_SUR} that have demonstrated excellent results. SINDy~\cite{SINDy} (Please refer to Tab.~\ref{tab:abb_table} for abbreviations), based on sparse regression, aims to extract the dynamical links within data. The main idea is to select a few optimal candidate functions from pre-defined over-complete dictionaries, which could be defined based on prior knowledge about the physical laws governing the process and effectively describe the dynamics. This is achieved by using a sparse regression method~\cite{LASSO, FOCUSS, SR3}. On the other hand, PDE Net, proposed by~\cite{PDENET}, aims to identify PDE equations by discretizing the PDE using forward Euler in time and finite difference in space. This method introduces a different approach mainly based on a succession of symbolic neural networks that aim to link multiple candidates using the EQuation Learning (EQL) architecture~\cite{EQL} and select the best ones using learnable convolutional kernels . Other methods use a different approach, SGA-PDE~\cite{SAIND} proposes the use of symbolic mathematics tree representation with classical and differential operators to represent the PDE and a genetic algorithm to find the best representation tree. In the same context, PySR~\cite{PYSR} introduces a Python library for symbolic regression based on multi-population evolutionary algorithm. This library is particularly well-suited for identifying ODEs, but has limited capabilities in identifying PDEs. This limitation arises from the fact that PySR's operators are restricted to basic algebraic operators. In contrast, SGA-PDE~\cite{SAIND} extends this capability by incorporating differential operators.
These methods offer high potential as the identified parameters can be transferred and used within traditional CAE simulation software finite elements. Additionally, they can be coupled with DNN such as PINN~\cite{PINN} to identify and simulate the dynamics simultaneously as proposed by~\cite{PINNSR}. \\

Nevertheless, these methods are not pertinent (inference time, adaptability, and leveraging new data to enhance accuracy...) in cases of online identification, where abrupt changes may occur during the process. In such scenarios, the model needs to rapidly adjust to these changes. SINDy-Model Predictive Control (MPC)~\cite{SINDYMPC} demonstrates that a combination of SINDy with control~\cite{SINDYc}, built on classical SINDy and DMDc~\cite{DMDc}, allows for better control of the system dynamics after a change. This method compared SINDYc, DMDc, and a DNN, revealing that the combination of SINDYc with DMDc yields interesting results. Moreover~\cite{ADAPTSINDy} suggests an adaptive model for abrupt changes based on SINDy. This method involves two processes: the first detects the changes, while the second identifies and then modifies, adds, or removes parameters from the dynamic system.\\
These methods present significant potential but have limitations when the identification time is crucial. A slight delay in identification can lead to substantial changes, such as in process systems prognosis~\cite{OASISP}. Developing an accurate dynamics identification model using limited data is not viable and may take a considerably long time, hindering its use for interactive applications. Therefore, using SINDYc or abrupt-SINDy online for real applications is challenging. \\

Recently, \cite{SEQ2SEQTR} presented a sequence to sequence Transformer~\cite{ATTTRF} architecture to train a general model to perform symbolic regression identification, with faster inference time. This technique is then generalized to ODE identifications, as presented in ODEFormer~\cite{ODEFormer} and NSODE~\cite{ODEFormer_u}. They introduced a transformer-based method to identify ODEs structure and parameters, utilizing Symbolic Regression techniques. While the first method targets univariate ODEs, the second extends to multivariate ODEs. These methods focus on training a general model to conduct forward identification of symbolic representations of ODEs based on time series observations. ODEFormer~\cite{ODEFormer} exhibits promising results compared to alternative methods. However, this approach is limited when a partial or full knowledge of the dynamics terms are available. Also these methods necessitate an embedding step and an additional fine-tuning step to refine the identified ODE parameters, posing challenges for real-world applications. Hence, the high complexity of the architecture may pose limitations, especially with long time series. Finally, since the model output uses natural language, it would be challenging to couple these methods with other deep learning methods, such as PINNSR~\cite{PINNSR}. \\
In the same context, OASIS~\cite{OASIS} has presented a framework that integrates both the strengths of SINDy for identifying the dynamics parameters and a DNN trained offline with parameters identified by SINDy. The latter is subsequently used online to provide parameter predictions for predictive control. OASIS has also been applied for fault real-time prognosis and demonstrated interesting and accurate results~\cite{OASISP}. The trained neural network uses a single time frame input which cannot handle information about sequence data. Additionally, classical neural network architectures may fail to encode the interactions between data, and therefore struggle to capture the dynamical system, especially with complex systems.\\

\subsection{Paper contributions}
We introduce a deep learning architecture for evolving online dynamics identification based on Set Encoding. Our goal is to enhance the encoding of interactions within data, leading to improved prediction accuracy of system parameters. We employ two distinct architectures Deep Set~\cite{DEEPSETS} and Set Transformer~\cite{SETTRANS}. 
These methods use two different encoding techniques, that can learn both simple pairwise relationships and higher-order links between set data, which can be applied to diverse datasets, making it suitable for a wide range of applications including supervised (our work) and unsupervised learning. 
In this work, we are particularly interested in applying this technique to sequential data, notably time series. Our main focus is to combine these methods with SINDy to provide an identification framework of the dynamics parameters of a fixed ODE for online use cases such as time series forecasting. \\
The problem addressed in this article is defined in Eq.~\ref{eq:prblstatment}:

\begin{equation}
\label{eq:prblstatment}
\begin{aligned}\left\{\begin{matrix}
\frac{\mathrm{d} X}{\mathrm{d} t} = \Theta (X) ~\Xi ~~~ \textup{where} ~~~ X~=~[x_0,~\dots,~x_n]\\
\Theta (X)~=~[1,~X,~X^2,~\dots,~X^2\otimes\sin(X),~\cos(X),~\dots] ,\\
\mathbf{G}(\Theta (X)) = \Xi
\end{matrix}\right.\end{aligned}
\end{equation}

\hspace{-15pt}where $\Theta(X)$ is a pre-defined search library based on knowledge about the dynamical system, $X$ is a vector containing the variables of the dynamical system, $\Xi$ represents the parameters of the dynamical system to be calculated. The main goal is to train the Set Encoding methods represented as $\mathbf{G}$ to predict the $\Xi$ matrix, thereby identifying the dynamics system.

\hspace{-15pt}The main contributions of the study are to : 
\begin{itemize}
\item Propose an appropriate approach to encode and decode to effectively learn the interactions between data, which provide accurate predictions of dynamics parameters. This architecture can handle sequential data with variable sizes, offering enhanced predictions compared to single-time point predictions. 
\item Illustrate that integrating physics loss in offline learning contributes to a better convergence and generalization of the model. 
\item Demonstrate that these methods can provide robust predictions with data containing  noise.
\item Provide accurate predictions (after offline training) with an inference time of milliseconds, making it suitable for real-time applications.
\end{itemize} 

This article is structured as follows: the section~\ref{SEC2and3} will be dedicated to a brief presentation of related works such as SINDYc which will be used in the following sections, and the OASIS framework along with its training method. Then, in section~\ref{SEC4}, we introduce the two set encoding architectures and discuss their use for online dynamics identification. Section~\ref{SEC5} will focus on comparing OASIS and set encoding for local dynamics identification using the Lotka-Volterra system, followed by a comparison of the two set encoding architectures for global dynamics identification using the Lorenz system, \textcolor{black}{then using Deep Set for 1D heat-transfer identification and characterization}. Finally, a conclusion of this work and areas for improvement are given.\\\\


\section{Related works}\label{SEC2and3}
In this section, we briefly introduce SINDYc (Sec.~\ref{SEC2}) and OASIS (Sec.~\ref{SEC3}). SINDYc is utilized to generate training data for comparing OASIS and our methods in the first application with the Lotka-Volterra system (Sec.~\ref{application1}), as well as for comparing the two Set Encoding methods presented in Sec.\ref{SEC4} for the second application with the Lorenz system
 (Sec.~\ref{application2}).
\subsection{SINDYc}\label{SEC2}
SINDYc~\cite{SINDYc} method allows the identification of parameters of a dynamical system with control variables. Drawing upon prior knowledge of the system's dynamic behavior, a set of candidate functions, referred to as the search space or library denoted by $\Theta$, is defined. These functions can collectively represent the dynamical system. The underlying idea is primarily based on the notion that an identified system should encompass a limited and concise set of functions. This approach can be applied to various dynamic systems (ODE and PDE), as well as to dynamic systems with control variables (Eq.~\ref{CDS}), which are external variables. 

\begin{equation}\label{CDS}
\dot{X}=f(X, U),~~~~X(t_0)=X_0
\end{equation}

\hspace{-15pt}Where $X=[x_1~~x_2 ~~ \cdots ~~ x_l]\in\mathbb{R}^{n\times l}$ is a matrix with $l$ state variables with $n$ time snapshots, $\dot{X}=[\dot{x_1}~~\dot{x_2} ~~ \cdots ~~ \dot{x_l}]\in\mathbb{R}^{n\times l}$ is the time derivative of $X$, usually approximated using total variation regularized derivatives~\cite{chartrand2011numerical}, and $U=[u_1~~u_2~~\cdots~~u_p]\in\mathbb{R}^{n\times p}$ are the $p$ control variables with $n$ time snapshots, and $X(t_0)$ are the initial conditions. The equation identified by SINDYc is defined as :

\begin{equation}\label{SINDYc}
\dot{X}~~=~~\Theta(X, U)\Xi ,
\end{equation}

\hspace{-15pt}whereas $\Xi$ is a coefficient matrix representing the selected functions from the search library $\Theta(X,U)$, which is defined by multiple functions (Eq.~\ref{theta}) usually using partial knowledge of the physics, for example, the library function could be defined as:

\begin{equation}\label{theta}
\Theta(X,U) = [1~~X^T~~U^T~~(X\otimes U)^T~~\cdots~~\sin(X^T)~~\cdots~~U^T\otimes \exp(X^T)~~\cdots]
\end{equation} 

To identify the dynamical system while avoiding overfitting, the $\Xi$ matrix must be sparse, implying that the less significant coefficients need to be equal to zero. This is achieved by solving the LASSO optimization problem (Eq.~\ref{optimsindyc}) defined  as : 
\begin{equation}\label{optimsindyc}
\Xi = \underset{\widetilde{\Xi}}{\mathrm{arg min}}  \left [  \left\|  \dot{X} - \Theta(X, U)\widetilde{\Xi}\right\|_2+\lambda\left\| \widetilde{\Xi} \right\|_1 \right ], 
\end{equation} 

\hspace{-15pt}where the first term represents the residual error and the second term is a regularizer applied to the coefficients matrix $\widetilde{\Xi}$ or the parameters of the ODE/PDE, and $\lambda$ is a weight parameter that imposes regularization. The larger it is, the stronger the regularization becomes, leading to the elimination of more terms from the search library while potentially reducing the accuracy. Other methods could be used such as Sparse Relaxed Regularized Regression (SR3)~\cite{SR3}, or thresholded least squares that use the norm 0 to eliminate less significant coefficients~\cite{SINDy, STRIdge, FOCUSS}.

Despite all the advantages of SINDYc, this method is restricted in cases where system parameters tend to vary during the process, or whenever the system must be identified online based on sensor data. While abrupt-SINDy\cite{ADAPTSINDy} presents a method to address this issue, it has drawbacks, notably the quantity of data required for identification and the computational cost, which can be significant for real-time cases. \textcolor{black}{In this study we will focus on using SINDy and SINDYc as offline identification methods to generate some datasets, which are then used to train Deep Neural Networks to perform the approximation of parameters online}.

\subsection{OASIS}\label{SEC3}
\begin{figure}[!h]
    \centering
    \includegraphics[width=0.9\linewidth]{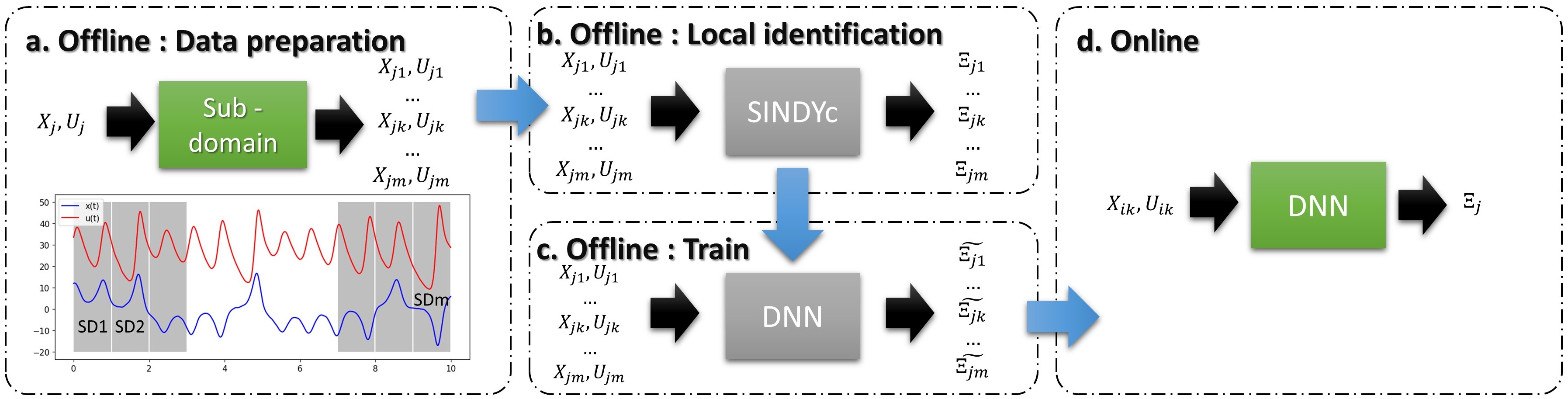}
    \caption{OASIS training process. $X_{jk}\in\mathbb{R}^{N^{'}\times l}$ and $U_{jk}\in\mathbb{R}^{N^{'}\times p}$ represent respectively state variables and control variables for the $j$-th set and the $k$-th sub domain, where $j \in [1, S]$ and $k \in [1, m]$. Here, $S$ is the number of sets and $m$ is the total number of sub domains. $l$, $p$, and $N^{'}$ represent respectively the number of state variables, control variables, and time windows.}
    \label{fig:OASIS_FR}
\end{figure}

The OASIS framework is primarily based on SINDYc but unlike the latter, it favors the use of offline trained DNN  through results from SINDYc, making it more suitable for real-time parameter identification cases.\\
The dynamical system (Eq.~\ref{CDS}) is a set of measured or generated time series. Then, the set is divided into multiple sub domains (Fig.~\ref{fig:OASIS_FR}.~a), which are further used for offline identification through SINDYc (Eq.~\ref{SINDYc}) using an identical search library (Fig.~\ref{fig:OASIS_FR}.~b).

\begin{figure}[!h]
    \centering
    \includegraphics[width=0.7\linewidth]{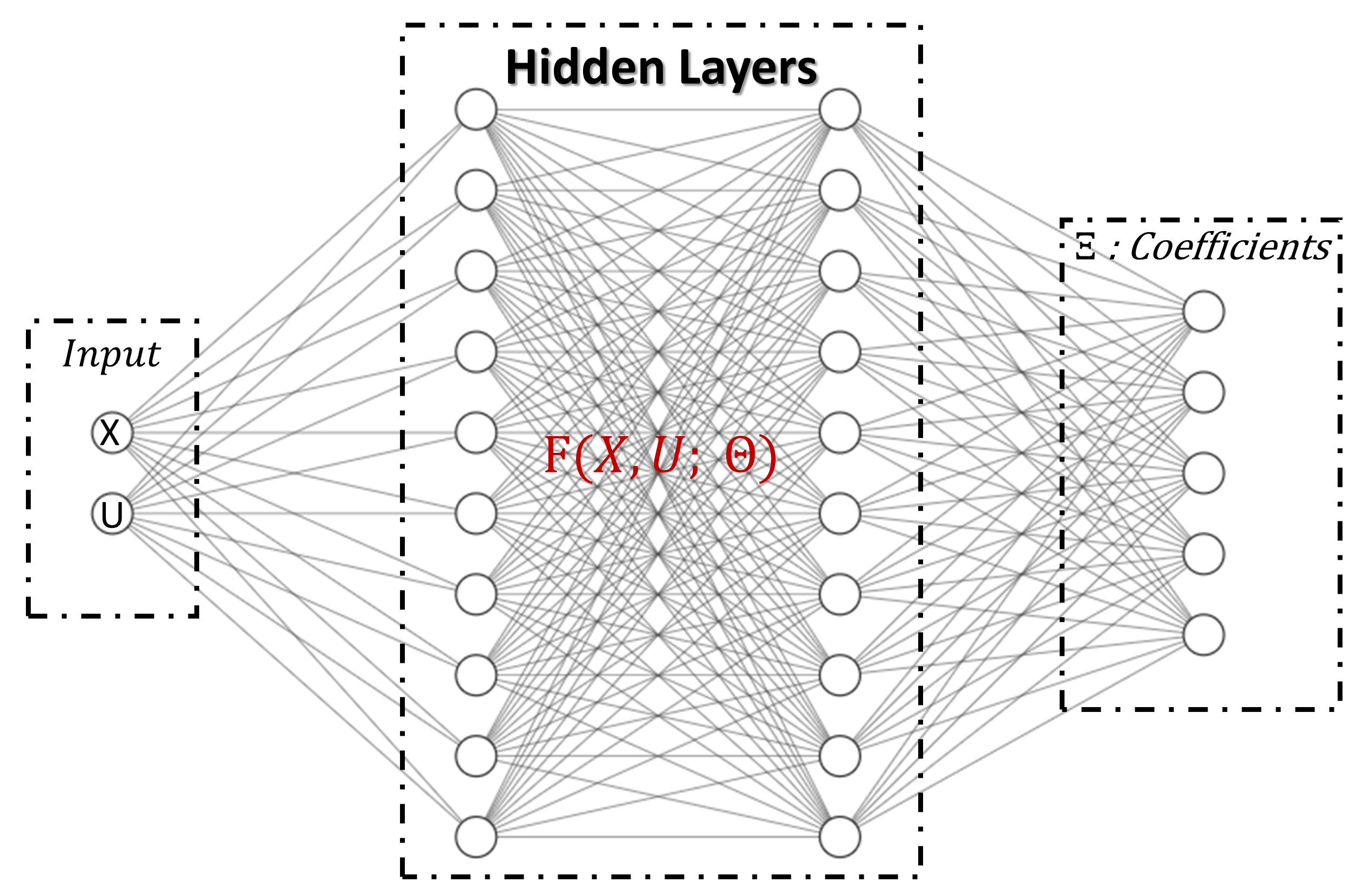}
    \caption{Deep neural network with two hidden layers with two inputs denoted as $X$ and $U$, and five outputs denoted as $\Xi$. The function $F$, which is trainable, depends on a set of variables denoted as $\Theta$ that governs the connections between these hidden layers.}
    \label{fig:OASIS_DNN}
\end{figure}

This dataset is then utilized for training the DNN (Fig.~\ref{CDS}) model (Fig.~\ref{fig:OASIS_FR}.~c), where the input comprises state and control variables and the output corresponds to the predicted $\Xi$. The training process involves leveraging the information contained in the datasets, enabling the model to learn and capture the underlying patterns and relationships within the data, allowing accurate and online identification (Fig.~\ref{fig:OASIS_FR}.~d). 
This is achieved by minimizing the error between the predictions of the DNN and the true coefficients (Eq.~\ref{eq:DNN}). 
\textcolor{black}{\begin{equation}
\label{eq:DNN}
\underset{\Theta}{\mathrm{arg min}} ~~ \left [  \mathrm{F}(X,U; ~\Theta) - \Xi + \beta \left\| \Theta \right\|_2 \right ] ~~~\mathrm{where}~~~ \mathrm{F}(X,U; ~\Theta) = \hat{\Xi}
\end{equation}}
Here, $\mathrm{F}$ is the output prediction of the DNN (Fig.~\ref{fig:OASIS_DNN}), and $\Theta$ denotes its trainable parameters. Additionally, it is important to include regularization terms $\beta \left\| \Theta \right\|_2$ to mitigate overfitting problems, where $\Theta$ represents the trainable variables of the DNN.

The trained model is then employed in real-time for online identification of the dynamical system~ (Eq.~\ref{fig:OASIS_DNN}) coefficients $\Xi$, enabling accurate extrapolation or prediction of optimal control parameters.\\

Because the DNN inputs are the current states of the system $X$  and controller $U$,  the OASIS framework does not aim to identify the global dynamics; this means the primary goal is not to identify the true and global coefficients of the governing ODE, but rather a local dynamic system limited to a small time range, and that provides limited extrapolation or prediction of optimal control parameters. This approach is motivated by the fact that accurate identification of the global dynamics is not straightforward and not always feasible~\cite{OASISP}.\\
In seeking to identify a valid local dynamics within a limited time interval, frequent model updates are necessary, which can lead to computational issues. To address this concern, the strategy of OASIS-P~\cite{OASISP} aims at choosing an optimal frequency for model updating, minimizing the identification error and consequently the extrapolation error.\\

\begin{figure}[!h]
    \centering
    \includegraphics[width=0.9\linewidth]{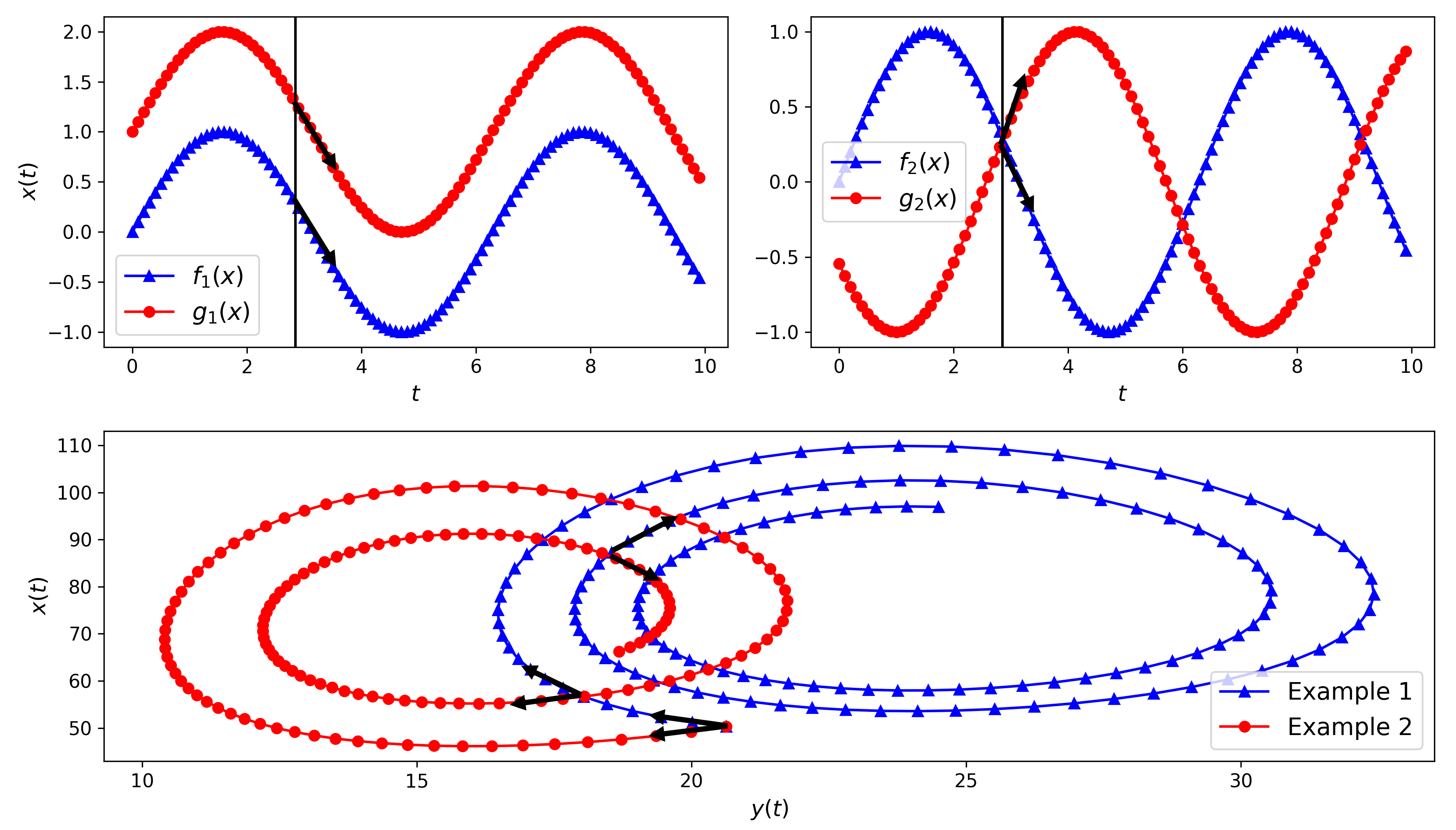}
    \caption{In top : two examples showcase the dynamics where the model can easily capture the local behavior with only one point in time (\textcolor{black}{left image}) and another one where the model will fail to capture the local behavior due to confusion (\textcolor{black}{image at the right}). In the bottom : example of dynamics with two state variables system (Lotka–Volterra) where the model could fail to capture the local dynamics at certain points.}
    \label{fig:ISSUEEXALL}
\end{figure}

The fact that the networks use only a single state to predict the parameters defining the local system behaviors poses a potential restriction whenever the dynamics cannot be captured from a single time frame but require a longer observation window. This approach is expected to yield accurate predictions (\textcolor{black}{example in Fig.~\ref{fig:ISSUEEXALL}~top~left}). However, the limitation is attributed to the non-uniqueness of the solution with a single state, as demonstrated by the example in \textcolor{black}{Fig.~\ref{fig:ISSUEEXALL}~top~right} where the intersection of the two time series may introduce confusion during the model training process. \textcolor{black}{The Fig.~\ref{fig:ISSUEEXALL}~bottom presents two examples of the Lokta-Volterra dynamics with similar initial conditions and control variables but with small changes in the dynamics parameters. The black arrows show cases where the model inputs are similar ($x(t_i)$ and $y(t_i)$) for the two examples (example 1 in blue and example 2 in red), but the dynamics differ}. This aligns with the model's inadequate capacity to capture temporal information effectively, resulting in less accurate predictions.
The following section will present a more suitable architecture capable of effectively encoding temporal information, thereby providing better predictions.

\section{Invariant Deep Learning model for Nonlinear Dynamics Identification}\label{SEC4}

\begin{figure}[ht]
    \centering
    \includegraphics[width=0.7\linewidth]{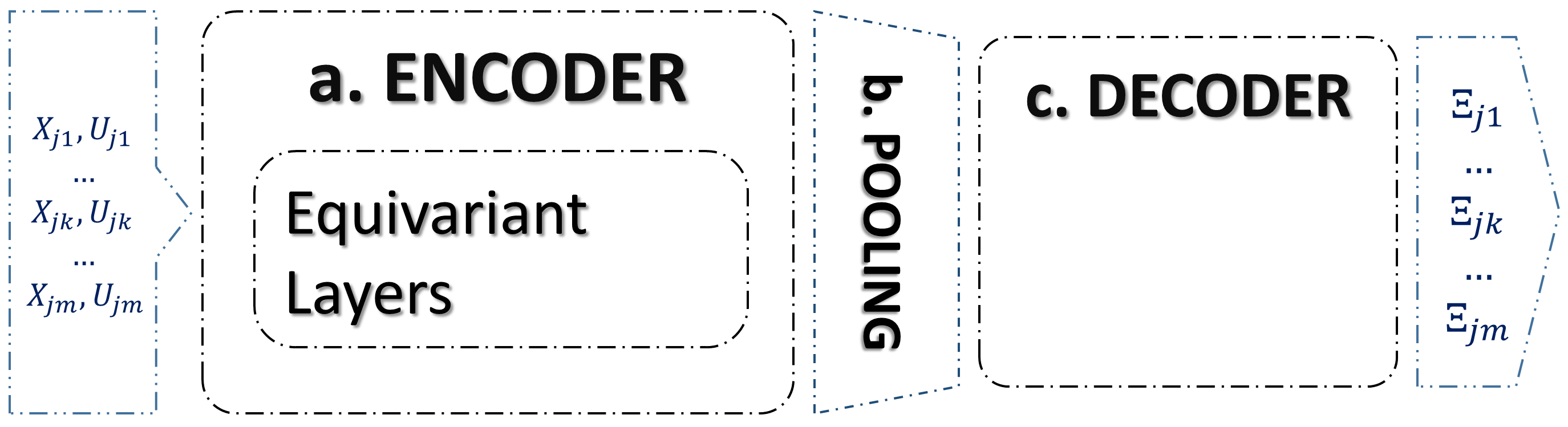}
    \caption{Set encoding overall architecture : The input sets, $X_{jk}$ and $U_{jk}$, represent respectively state variables and control variables for the $j$-th set and the $k$-th sub domain, that corresponds to the output coefficient $\Xi_{jk}$, $j \in [1, N]$ and $k \in [1, m]$. $N$ is the number of sets and $m$ is the total number of sub domains.}
    \label{fig:setenco}
\end{figure}

In this section the proposed architecture is founded on a concept termed Set Encoding. In contrast to classical neural network models (Fig.~\ref{fig:OASIS_DNN}), where the primary goal is to learn a function $F$ connecting a fixed-dimensional input tensor to its corresponding output value, set encoding (Fig.~\ref{fig:setenco}) could be used to establish links between input data sets with variable shapes (variable time series length and variable time steps $\Delta t$) and their corresponding labels, while maintaining permutation invariance. This objective is realized by encoding input data sets through equivariant layers (Fig.~\ref{fig:setenco}~a), followed by pooling steps to attain permutation-invariant properties (Fig.~\ref{fig:setenco}~b). Subsequently, the information is decoded (Fig.~\ref{fig:setenco}~c) to yield the appropriate outputs.\\

In this section, we explore the applicability of two set encoding methods for online and adaptative dynamics identification, utilizing variable shapes of time series. The first method, Deep Set, is based on a DNN (Fig.~\ref{fig:OASIS_DNN}) architecture, while the second method, Set Transformer, is built upon a transformers architecture~\cite{ATTTRF}.

\subsection{Deep Set}\label{DeepSet}
\begin{figure}[!h]
    \centering
    \includegraphics[width=0.7\linewidth]{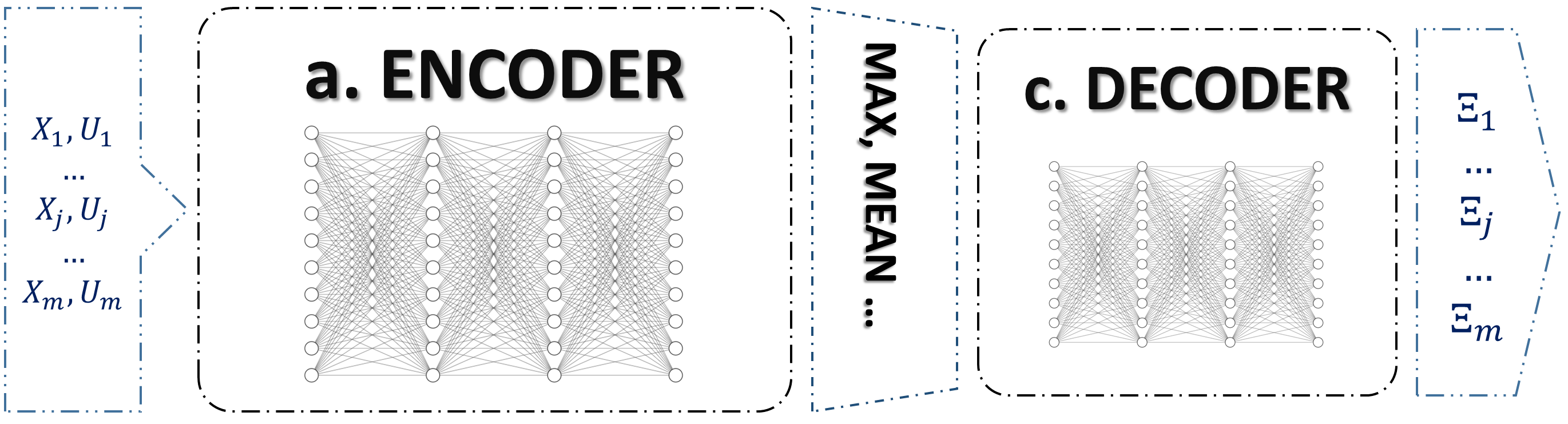}
    \caption{Deep Set Architecture : the encoder is a DNN followed by a pooling transformation (max, mean, min ...),  which is then decoded with another DNN.}
    \label{fig:deepset}
\end{figure}

The Deep Set architecture (Fig.~\ref{fig:deepset}) can be divided into three transformations. The first step involves encoding the inputs into a latent space through a series of layers with non-linear activation functions (Eq.~\ref{eq:deepsetencdec}). In our cases the input is a set of $X_{jk}\in\mathbb{R}^{N^{'}\times l} $ and $U_{jk}\in\mathbb{U}^{N^{'}\times p}$, where $l$, $p$, and $N^{'}$ represent respectively the number of state variables, control variables, and the time series length.
This transformation can be made permutation-equivariant by employing suitable layers. This implies that the output remains equivariant to permutations of set elements. This objective can be accomplished by defining the neural network layers as in Eq.~\ref{eq:equivariant}, as demonstrated by~\cite{DEEPSETS}.
\begin{equation}
\label{eq:equivariant}
\begin{aligned}\left\{\begin{matrix}
\mathbf{f(x)} \doteq  \mathbf{\sigma(\lambda I x+\gamma pool(x))} \\
\sigma :~~~\textup{nonlinear activation function}\\
\lambda,~\gamma  :~~~\textup{learnable variables}\\
\mathbf{pool(x)} :~~~ \textup{max, sum, mean ...}
\end{matrix}\right.\end{aligned}
\end{equation}

Following this encoding, a pooling transformation is applied (such as max, mean, sum, etc.) with respect to the time dimension which yields a set of latent variables  $Z\in\mathbb{R}^{F\times L}$, as defined in Eq.~\ref{eq:deepsetencdec}  :
\begin{equation}
\label{eq:deepsetencdec}
\begin{aligned}\left\{\begin{matrix}
\textbf{Encoder}(X_{jk}, U_{jk}) &= &\textbf{Pool}(\textbf{DNN}(\left\{   X_{jk}(t_0),~\ldots,~ X_{jk}(t_{N^{'}}),~ U_{jk}(t_0),~ \ldots,~ U_{jk}(t_{N^{'}})  \right\})) = Z\\
\textbf{Decoder}(Z) &=& \textbf{DNN}(Z) = \Xi_{jk} ,
\end{matrix}\right.\end{aligned}
\end{equation}

\hspace{-15pt}where $F$ in $Z\in\mathbb{R}^{F\times L}$ represents the number of encoded features. This operation enables the model to manage inputs of variable shapes, thereby achieving an invariant-type (Eq.~\ref{eq:invariant}) encoding.
\begin{equation}
\label{eq:invariant}
\begin{aligned}
f(\{x_1,\cdots ,x_n\}) = f(\{x_{\pi(x_1)},\cdots ,x_{\pi(x_n)}\}) ~~~~& \textup{for any permutation }\pi
\end{aligned}
\end{equation}

\hspace{-15pt}The final step involves decoding the latent vector to obtain the desired output. In this application, we will only utilize the invariant property. 

\subsection{Set Transformer}\label{SetTransformer}
\begin{figure}[!h]
    \centering
    \includegraphics[width=0.9\linewidth]{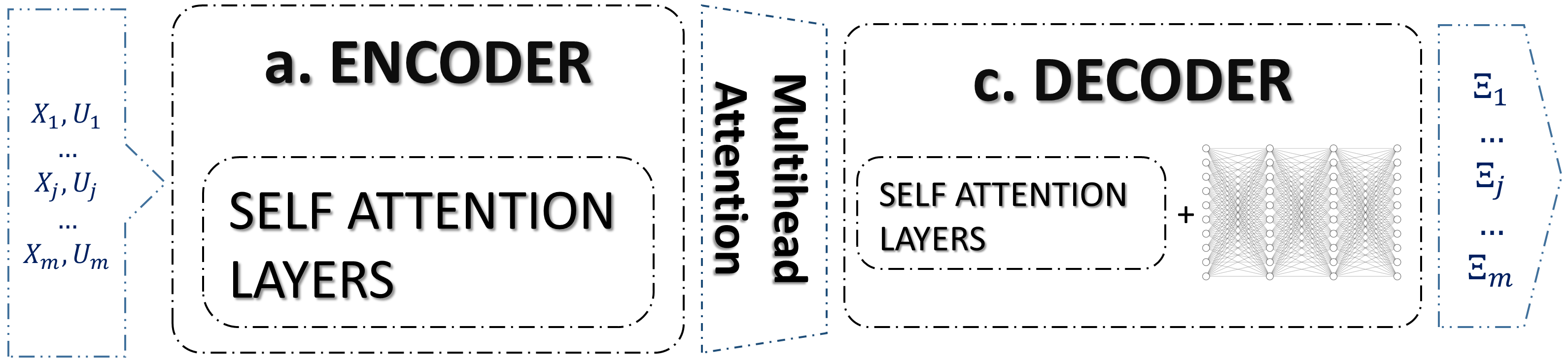}
    \caption{Set Transformer Architecture: the encoder is a succession of self-attention layers followed by a pooling layer, which is then decoded with self-attention layers and a DNN.}
    \label{fig:settrans}
\end{figure}

The architecture of set transformer (Fig.~\ref{fig:settrans}), is based on transformer model~\cite{ATTTRF}. The motivation behind the use of this architecture lies in its ability to encode links and pairwise interactions among the elements of sets through the attention mechanism. The architecture is primarily built on layers called self attention layers, constructed to be permutation-equivariant.\\

\hspace{-15pt}The attention mechanism can be defined by the Eq.~\ref{eq:attmecha} :
\begin{equation}
\label{eq:attmecha}
\mathbf{Att}(Q,K,~V;\sigma)=\sigma(QK^T)V. 
\end{equation}
Where $Q$ is a matrix of $n$ sets with the same dimension $d_q$, and $K$ and $V$, referred to as Keys and Values, are two matrices of sets with the same number of sets. The dimension of $K$  is the same as that of $Q$, while $V$ has a different dimension $d_v$. The initial part of the equation, $\sigma(QK^T)$, measures the similarity between the two sets of matrices, where $\sigma$ usually denotes the softmax function applied to the product to convert it into a probability distribution. The result of this product represents a weight matrix, which is used to assign different importance to the elements of the matrix $V$.\\

Utilizing this mechanism empowers the model to capture relevant patterns and dependencies in the data. To achieve this objective, the mechanism is extended to Multi-head Attention (Eq.~\ref{eq:multiattmecha}) defined by three transformations. 
\begin{equation}
\label{eq:multiattmecha}
\begin{aligned}\left\{\begin{matrix}
\mathbf{MultiHead}(Q,K,V;\sigma) = \mathrm{concatenation}(O_1,\cdots,O_j,\cdots,O_L)W^o\\
O_j = \mathbf{Att}(QW^q_j,KW^k_j,VW^v_j;\sigma_j)
\end{matrix}\right.\end{aligned}
\end{equation}

The initial step involves projecting the $Q$, $K$, and $V$ matrices to a higher dimension $d_q^H=d_k^H$ and $d_v^H$ using learnable parameters ${W^q_j, W^k_j\in\mathbb{R}^{d_k\times d_k^H},~~ W^v_j\in\mathbb{R}^{d_v\times d_v^H}}$. Subsequently, the attention mechanism is applied to the $L$ resulting projections, referred to as Head Attentions $O_j$. These Head Attentions are then concatenated and multiplied by a trainable variable matrix $W^o\in\mathbb{R}^{Ld_v^H\times d}$. In this application, it was considered that $d_q=d_v=d$, and $d_q^H=d_v^H=d/L$.
The Set Transformer proposes set operation blocks, utilizing a Multi-Head Attention mechanism to encode the data and subsequently aggregate and decode the encoded features. The data is encoded through Multihead Attention Blocks (MAB) and Set-Attention Blocks (SAB). Given two input sets matrices $X, Y \in \mathbb{R}^{n \times d}$, MAB is defined as: \\

\begin{equation}
\label{eq:MAB}
\begin{aligned}\left\{\begin{matrix}
\mathbf{MAB}(X,Y) =\mathbf{Normalization}(H+\mathbf{rFF}(H)), \\
H=\mathbf{Normalization}(X+\mathbf{MultiHead}(X,Y,Y)).
\end{matrix}\right.\end{aligned}
\end{equation}

The $\mathbf{Normalization}$ is the layer normalization~\cite{NORMLAYER}, and $\mathbf{rFF}$ represents a row-wise feed forward neural network without activation functions. The SAB block is defined as :
\begin{equation}
\label{eq:SAB}
\mathbf{SAB}(X)=\mathbf{MAB}(X,X)
\end{equation}
Where the Keys and Queries are identical, employing the self-attention mechanism to capture interactions between input features. 
The encoding is then defined by a succession of SAB blocks (Eq.~\ref{eq:STRENCODAGE}).
\begin{equation}
\label{eq:STRENCODAGE}
\mathbf{ENCODER}(X)=\mathbf{SAB}(\mathbf{SAB}(X))
\end{equation}
The output of the latter is determined by the function $F: X \to Z \in \mathbb{R}^{n \times d}$, where $X \in \mathbb{R}^{n \times d_x}$. This output is subsequently aggregated through the PMA (Pooling Multihead Attention) block defined as :

\begin{equation}
\label{eq:STRDECODAGE}
\begin{aligned}\left\{\begin{matrix}
\mathbf{DECODER}(Z) =\mathbf{rFF}(\mathbf{SAB}(\mathbf{PMA}(Z))), \\
\mathbf{PMA}(Z)=\mathbf{MAB}(S, \mathbf{rFF}(Z)).
\end{matrix}\right.\end{aligned}
\end{equation}

\hspace{-15pt}Where $S \in \mathbb{R}^{k \times d}$ is a trainable matrix, with $k$ being the desired output features (1 in our case). This allows for an aggregation based on the importance of set elements rather than equal importance (as in the cases of max, mean, etc., aggregation). Following this transformation, a SAB is applied to better capture the connections between the $k$ output features (Eq.~\ref{eq:STRDECODAGE}).

Considering the quadratic time complexity of the Set Attention Block (SAB) as $O(n^2)$, where $n$ represents the number of sets, \cite{SETTRANS} introduced an improved version named Induced Set Attention Block (ISAB). The ISAB offers reduced computational complexity while preserving the same performance.\\\\

\textcolor{black}{The architectures of both models are more complex compared to OASIS, which may raise concerns, particularly regarding overfitting problems. We address these concerns in some examples in the next section. The selection of hyperparameters can be managed using classical testing and validation approaches, which we have employed in this work. In the following section, we demonstrate that this complexity offers significant advantages through three different applications. Additionally, we provide solutions and perspectives to mitigate potential overfitting issues.}

\section{Results}\label{SEC5}

\subsection{Tests cases}\label{SEC51}
To demonstrate the advantages of Set Encoding methods, we present \textcolor{blue}{three} test cases with \textcolor{blue}{three} different systems. In the section~\ref{application1} we present a comparison of OASIS framework and Set Transformer for online nonlinear dynamics identification. The Lotka–Volterra~\cite{wangersky1978lotka}~(Eq.~\ref{eq:lotka}) system is used to compare these two methods. 
\begin{equation}
\label{eq:lotka}
\left\{\begin{matrix}
\begin{aligned}
\dot{x} &= \alpha x - \beta xy \\
\dot{y} &= \delta xy - \gamma y + c
\end{aligned}
\end{matrix}\right.
\end{equation}

First, we generate a dataset with various initial conditions and control variables. We then divide the time domain into multiple sub domains, and SINDYc is subsequently employed to identify the local dynamics model for each sub domains. Both methods are then trained to capture the local dynamics using time series and their corresponding identified local SINDYc mode.\\

While OASIS methods are limited to online local system identification, Set Encoding methods could also be applied in global non-linear dynamics identification. In the section~\ref{application2} we compare the two Set Encoding. These methods can handle variable input sizes, which means capturing the true dynamics through a sequence of data is possible. Furthermore, it can capture the interactions between input data, all of which help enhance the identification process. To study the usefulness of these methods, Set Transformer and Deep Set are compared for the Lorenz system (Eq.~\ref{eq:lorenz}). 
\begin{equation}\label{eq:lorenz}
\left\{\begin{matrix}
\dot{x} &=& \multicolumn{1}{l}{\sigma(y(t) - x(t))}\\
\dot{y} &=& \multicolumn{1}{l}{\rho x(t) - y(t) - x(t)z(t)}\\
\dot{z} &=& \multicolumn{1}{l}{x(t)y(t) - \beta z(t)}
\end{matrix}\right.
\end{equation}
\\

\textcolor{black}{Monitoring and control of industrial processes are generally based on fixing the theoretical parameters of the process and material properties. However, these parameters can change during the process. Therefore, it is important to use sensor data to identify these parameters in real-time. If any deviations are detected, control systems can be used to correct the process by adjusting the parameters, thereby improving its performance.
To address these challenges, Set Encoders can be applied for industrial process monitoring. For example, in the thermo-stamping process, a thin composite plate is heated to make it less rigid before being formed using a specific mold. Using real-time sensor data, the Set Encoder can approximate the material properties during the heating step, which helps, for example, to ensure the temperature is homogeneous throughout the depth of the plate to avoid defects during forming. In this application, we begin with a preliminary study of 1D heating using synthetic data before applying it to real data of the heating step in the thermo-stamping process in future work.\\
In~\ref{App3}, we will study the 1D heat-transfer problem (PDE example) using a Deep Set model. We will train the model to approximate diffusivity and to characterize an abnormality defined as a local decrease in diffusivity, using noisy data. The equation is given by (Eq.~\ref{eq:pde1d}):
\begin{equation}\label{eq:pde1d}
\left\{\begin{matrix}
\dot{T} &=& \multicolumn{1}{l}{\frac{\partial (\alpha(z) \partial  T)}{\partial z^2}}\\
-\frac{\partial  T}{\partial z}|_{z=0} &=& \frac{q_0}{k}(T)\\
\frac{\partial  T}{\partial z}|_{z=L} &=& 0 \\
T(t=0, z) &=& T_0
\end{matrix}\right.
\end{equation}}

\subsection{Application 1 : local model identifications (OASIS vs Set Transformer)}\label{application1}

\subsubsection{Data generation}\label{app1dataset}
\begin{figure}[!htbp]
    \centering
    \includegraphics[width=0.5\linewidth]{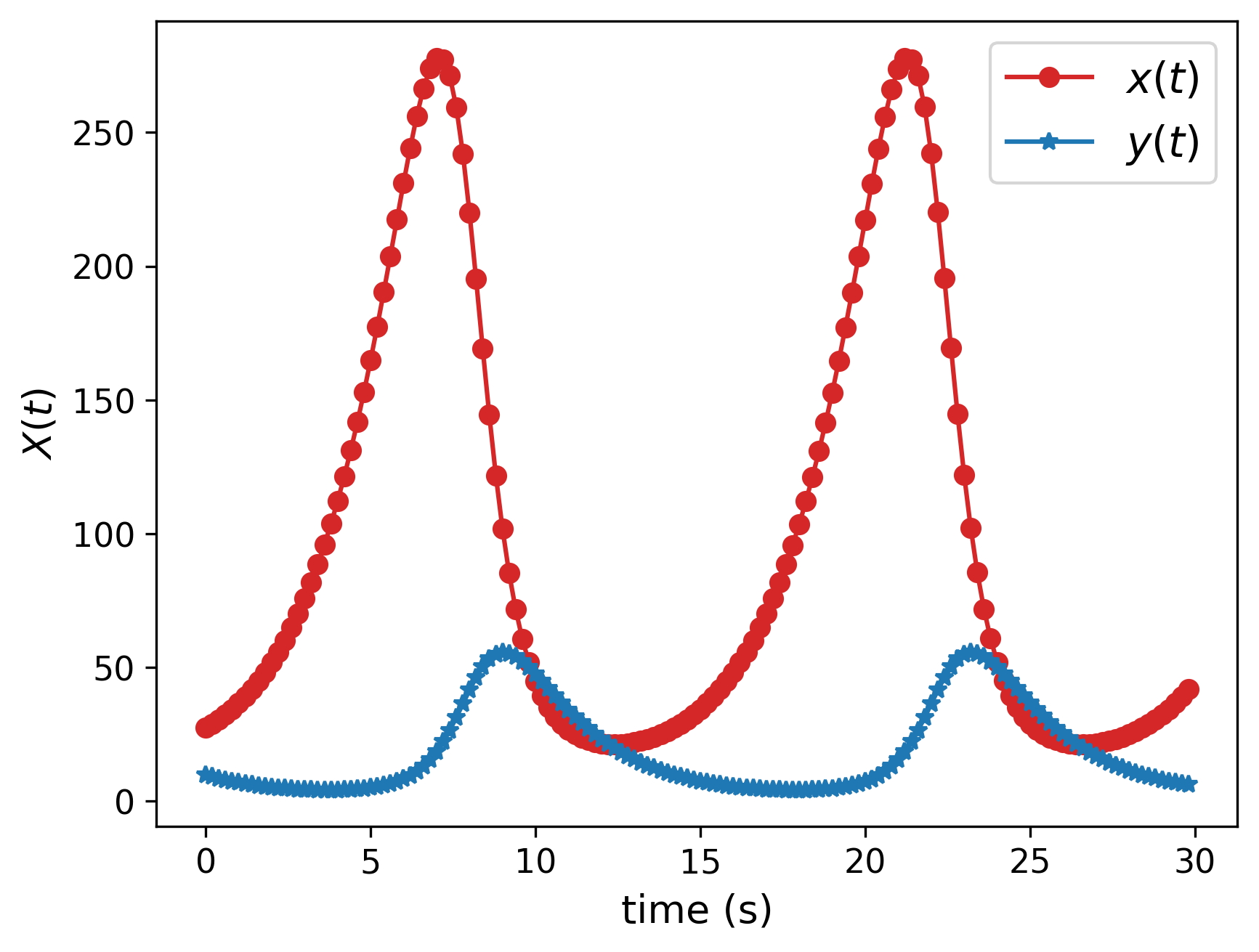}
    \caption{Example of Lotka–Volterra dynamics solution for \(\alpha = 0.5\), \(\beta = 0.025\), \(\delta = 0.5\), and \(\gamma = 0.005\) with a constant control variable $c=0$, $x(t_0)=27.5$ and $y(t_0)=10.0$ .}
    \label{fig:epp_eg}
\end{figure}

The Lotka–Volterra system (Eq.~\ref{eq:lotka}) is used to compare these two methods. Where, \(x\) and \(y\) respectively represent the populations of prey and predators. \(c\) is a control variable that limits the rate of increase or decrease of the predators. The parameters \(\alpha = 0.5\), \(\beta = 0.025\), \(\delta = 0.5\), and \(\gamma = 0.005\) represent respectively the rates of increase or decrease of prey and predators, as well as the effect of predators on prey (and vice versa). An example of a solution of Eq.~\ref{eq:lotka} is given in Fig.~\ref{fig:epp_eg}.

\begin{figure}[!h]
    \centering
    \includegraphics[width=1\linewidth]{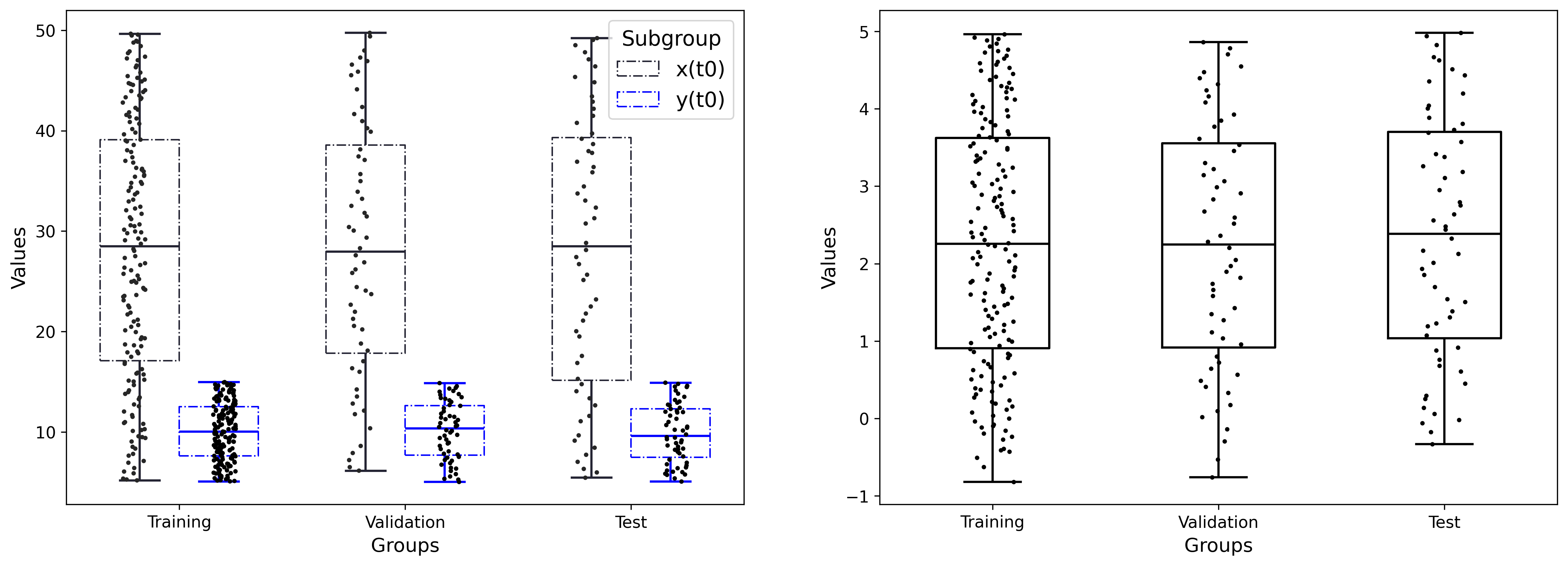}
    \caption{Left : initial value distribution $(x(t_0),~y(t_0))$ for training, validation, and test and Right : the control value distribution $c$ for training, validation, and test.}
    \label{fig:APP1DATADISTR}
\end{figure}

The dataset is generated through various combinations of control variable (\(c\)) and initial conditions \(x_0\) and \(y_0\), using the Sobol~\cite{SOBOL, LSODA} sampling method as given in Tab.~\ref{tab:res1DataSet}. The data generation is performed using the LSODA~\cite{LSODA} method from the SciPy Python package. In this comparison (Fig.~\ref{fig:APP1DATADISTR}), 278 datasets are generated, of which 166 will be used for model training, and 112 will be employed for model validation to mitigate overfitting issues, and a test set to compare the two methods. 

\begin{table}[!htbp]
\centering
\caption{Range of Variables }\label{tab:res1DataSet}
\begin{tabular}{@{}ll@{}}
\toprule
\textbf{Variable} & \textbf{Range} \\\midrule
\(c\) : Control variable & [-1; 5] \\
\(x_0\) : Initial condition $x$ & [5; 50] \\
\(y_0\) : Initial condition $y$ & [5; 15] \\
\(t\) : Time interval  & [0; 30] \\
\botrule
\end{tabular}
\end{table}

Each dataset is divided into several sub domains over time. SINDYc is then used to identify the local model of the different sub domains, which may be different from the global dynamics but can be used within a  short time interval. In this case, we assume that there is no initial information about the search library. The identification is performed using the PySINDy Python package~\cite{PYSINDY}. We use a base of polynomials of degree 3 defined as : 
\begin{equation}
\label{eq:theta}
\begin{aligned}
\frac{\mathrm{d}\widetilde{X}}{\mathrm{d} t} &= \Theta(\widetilde{X},C)~\Xi \\
\Theta(\widetilde{X},C) &= \begin{aligned}[b]
                  &[1,~X,~Y,~\ldots,~Y^3,~Y^2 C,~YC^2,~C^3] \in \mathbb{R}^{N\times 20}\hfill
                  \end{aligned},\\
                  \widetilde{X} &= \left [ X,~Y \right ]
\end{aligned}
\end{equation}
where $\Theta(\widetilde{X}, C)\in \mathbb{R}^{N\times 20}$ is the search library, where $C\in \mathbb{R}^{N\times 1}$ represents the control variables, and $\Xi \in \mathbb{R}^{2\times 20}$ is a sparse tensor representing the parameters to be fitted.

\subsubsection{Training process and Criteria for comparison}\label{app1compar}
Given the utility and significance of these methods in real-time control and defect detection, it is crucial to evaluate their extrapolation capabilities. For a fair comparison, both methods underwent training on the same dataset and were then assessed and validated using a separate portion designated for validation. The Tab.~\ref{tab:trainprocess} summarizes the training process for both methods. The error function is defined as a sum of three parts, the first one represents the difference between the true $\Xi$ and predicted $\hat{\Xi}$ parameters, the second part represents local dynamic loss by calculating the difference between the known derivative of $X$ and $Y$ and $\Theta(\widetilde{X},C)\hat{\Xi}$ where $\Theta(\widetilde{X},C)$ is know, the final term represents the regularization, contributing to the effective generalization of models. Both methods are trained using identical optimizer and a fixed number of epochs, incorporating variable learning rates during the training process. 
A noteworthy distinction pertains to the input of the two models. The first maps a single time step to $\hat{\Xi}$, whereas the Set Transformer encodes a temporal series of variable length. We also train the models to predict the dynamics of the subsequent sub domain based on the current sub domain's time series $\widetilde{X}_{t_i}$ and control variable $C_{t_i}$  as defined in Eq.~\ref{eq:models_predection} : 
\begin{equation}
\label{eq:models_predection}
F(\widetilde{X}_{t_i},C_{t_i})=\hat{\Xi}_{t_i+\Delta t} ~~~ \textbf{where} ~~~ \widetilde{X}_{t_i}\in \mathbb{R}^{N^{'}\times 2},~\widetilde{C}_{t_i}\in \mathbb{R}^{N^{'}\times 1},~\hat{\Xi}_{t_i+\Delta t}\in \mathbb{R}^{20\times 2}
\end{equation}
Where $N^{'}$ is the time snapshots, $N^{'}=1$ for OASIS framework and $N^{'}=\Delta t = 10$. Which will help in achieving better results for extrapolation.

\begin{table}[!htbp]
\centering
\renewcommand{\arraystretch}{1.5}
\caption{Training process summary}\label{tab:trainprocess}
\begin{tabular}{@{}llll@{}}
\toprule
                      & \multicolumn{1}{c|}{\textbf{Set Transformer}} & \textbf{OASIS} \\ \midrule
\textbf{Optimizer}        & \multicolumn{2}{c}{Adam}              \\ \midrule
\textbf{Loss function}           & \multicolumn{2}{c}{$\mathcal{L}_{\Xi} + \lambda_0 \mathcal{L}_{ode} + \lambda_1 \mathcal{L}_{reg} $=$\left \| \Xi -\hat{\Xi} \right \|_2 + \lambda_0\left \| \frac{d\widetilde{X}}{dt} - \Theta(\widetilde{X},C)\hat{\Xi} \right \|_1 + \lambda_1\left \| W \right \|_1$}              \\ \midrule
\textbf{Learning rate}           & \multicolumn{2}{c}{$\left [ 10^{-3},~4\times 10^{-4},~10^{-4},~10^{-5},~10^{-6},~10^{-7},~10^{-8} \right ]$}              \\ \midrule
\textbf{Epochs}           & \multicolumn{2}{c}{[2000,~2000,~4000,~2000,~2000,~2000,~2000]}              \\ \midrule
\textbf{Input Shape}  & \multicolumn{1}{c|}{$\left [Batch size,~time,~features \right ]$ : $\left [Batch size,~10,~4 \right ]$       }      &     $\left [Batch size,~1,~4 \right ]$        \\ \midrule
\textbf{Output Shape} & \multicolumn{2}{c}{$\left [Batch size,~coefficients \right ]$ : $\left [Batch size,~20 \right ]$}                            \\ \botrule
\end{tabular}
\end{table}

To ensure fairness and balance in the comparison, we used hyper-parameters tunning for training both methods, varying hyper-parameters like hidden layers, neurons per layer, regularization coefficient $\lambda_1$ , and activation functions. We then select the best architecture by assessing performance through validation error.

\subsubsection{Comparison and Conclusion}\label{application1compa}
The test set is not utilized during the training process; instead, the  validation portion is employed solely for selecting the best model during training. We also employ two models for each method to predict the coefficients of $\frac{d x}{dt}$ and $\frac{d y}{dt}$, which we believe will facilitate the training process.\\

\begin{figure}[!h]
    \centering
    \includegraphics[width=1.\linewidth]{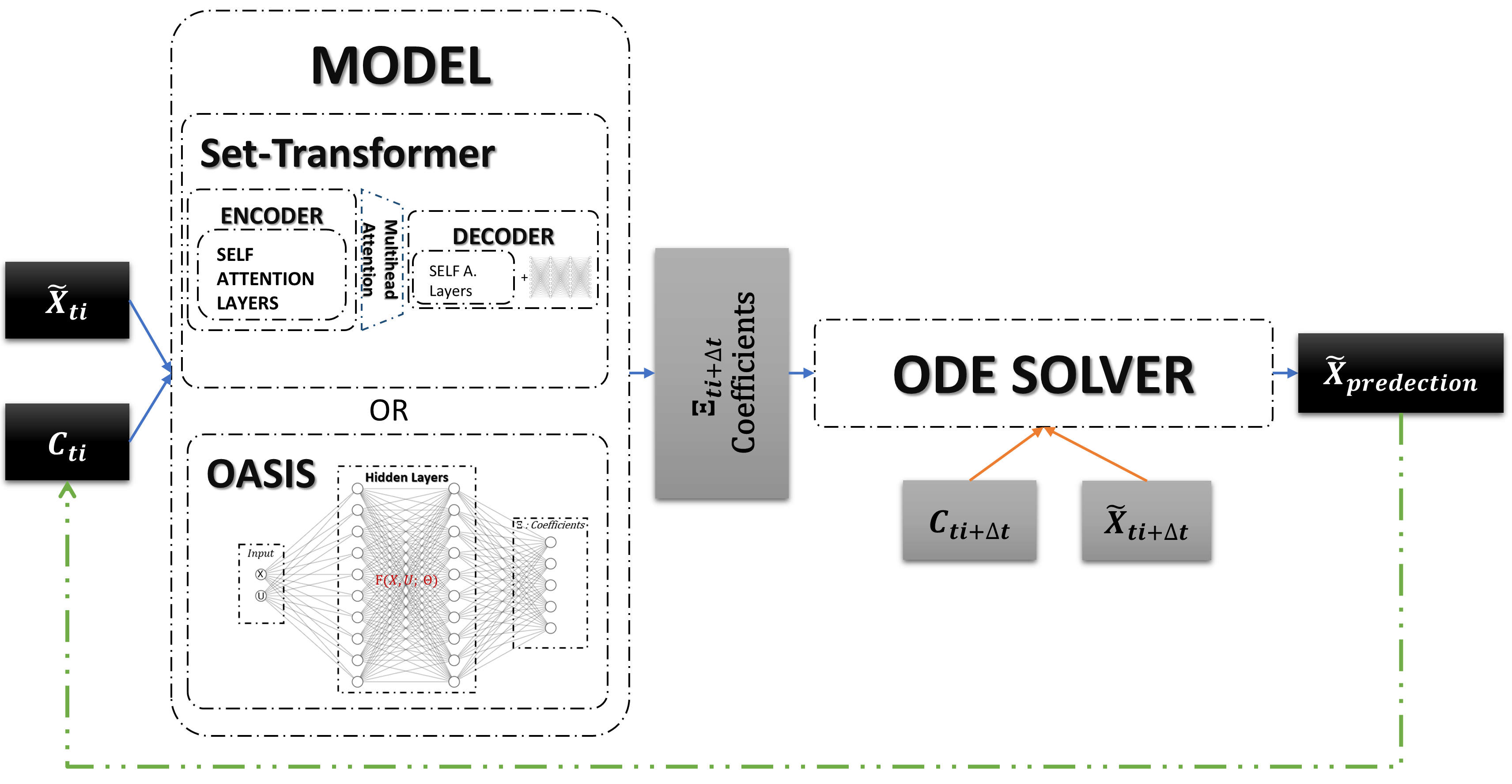}
    \caption{Comparison method of OASIS and Set Transformer : Extrapolation using local dynamic identification and ODE-Solver}
    \label{fig:compar_app1}
\end{figure}

\begin{figure}[h]
    \centering
    \includegraphics[width=1.05\linewidth]{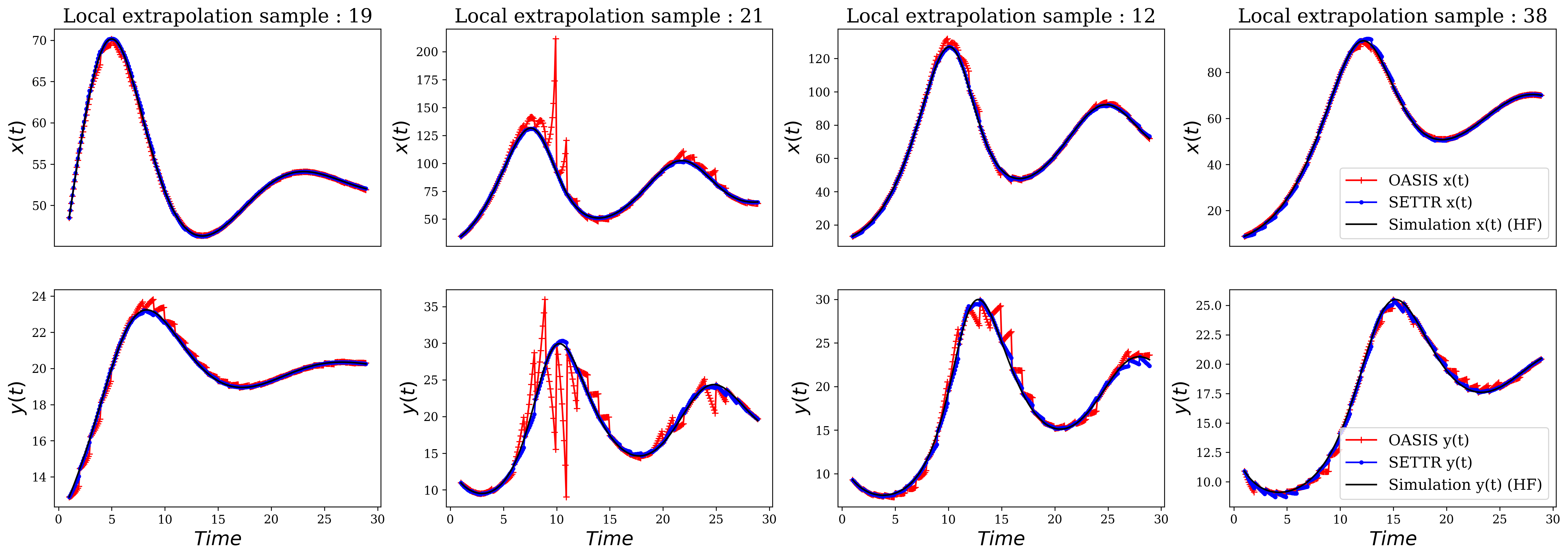}
    \caption{Example of extrapolation using Set Transformer in blue and OASIS in red compared to the ground truth in black (HF) simulation results.}
    \label{fig:app_eg_extrapolation}
\end{figure}

The models are then compared based on their ability to capture local dynamics and extrapolate (Fig.~\ref{fig:app_eg_extrapolation}) for a limited time range. This process is summarized in Fig.~\ref{fig:compar_app1}. The input of Set Transformer is a sequence, while the input of OASIS is a single point in time, first the models are used to identify the local dynamics $\hat{\Xi}^j_{t+\Delta t}$ which corresponds to the local dynamics of the following sub domain for the $j$ test dataset. It is then used with initial values $\widetilde{X}^j_{t_i+\Delta t}$ and control $C_{t_i+\Delta t}$ value to extrapolate using 'LSODA' as ODE-Solver, the output of the solver is a times series $\widetilde{X}^j_{prediction}=[x_j(t)_{p},~y_j(t)_p]~~\textup{where}~~t\in [t_i+\Delta t : t_i+2\Delta t]$, and $\Delta t$ corresponds to the time sub domain windows, which is then compared to the ground truth (simulation data) using MAPE (Eq.~\ref{eq:MAPE_error_x} and Eq.~\ref{eq:MAPE_error_y}).
\begin{equation}
\label{eq:MAPE_error_x}
\textbf{MAPE}_{t_i+\Delta t}^{x_j} =  \left\| 100\times\frac{x_j(t)_{p} - x_j(t)_{gt}}{x_j(t)_{gt}}  \right\|_1~~\textup{where}~~t\in [t_i+\Delta t : t_i+2\Delta t]~~\textup{and}~~j\in [1, N]
\end{equation}
\begin{equation}
\label{eq:MAPE_error_y}
\textbf{MAPE}_{t_i+\Delta t}^{y_j} =  \left\| 100\times\frac{y_j(t)_{p} - y_j(t)_{gt}}{y_j(t)_{gt}}  \right\|_1~~\textup{where}~~t\in [t_i+\Delta t : t_i+2\Delta t]~~\textup{and}~~j\in [1, N]
\end{equation}
This metric is usually used to measure the prediction accuracy. It quantifies the average percentage deviation between the predicted solution and the ground truth, providing a measure of how well the predictions align with the actual outcomes. Based on the analysis by~\cite{FORCASTBP} of time series forecasting evaluation, MAPE is a suitable metric in the absence of outliers and intermittence. We also utilize sMAPE, which is more suitable in the presence of outliers. This metric quantifies the percentage difference between predicted and actual values, with adjustments made to ensure equal treatment of both small and large values in the evaluation process (see Appendix~\ref{appendix_1}).

The two models are trained using fixed sub domain windows (sequences of 10 time points referred to as $\Delta t$), as explained in sections~\ref{app1compar} and ~\ref{app1dataset}, and then compared using the test dataset, with 3 different sub domain windows 10,~15 and 20, to evaluate the ability of the models to generalize and to capture the local dynamics.\\
Applying the same logic as explained earlier, we compute the Total MAPE for each dataset (Eq.~\ref{eq:OA_MAPE_error_x}) and Eq.~\ref{eq:OA_MAPE_error_y}). 

\begin{equation}
\label{eq:OA_MAPE_error_x}
\textbf{MAPE}_{Total}^{x_j} =  \left\| 100\times\frac{x_j(t)_{p} - x_j(t)_{gt}}{x_j(t)_{gt}}  \right\|_1~~\textup{where}~~t\in [\Delta t : T_f]~~\textup{and}~~j\in [1, N]
\end{equation}

\begin{equation}
\label{eq:OA_MAPE_error_y}
\textbf{MAPE}_{Total}^{y_j} =  \left\| 100\times\frac{y_j(t)_{p} - y_j(t)_{gt}}{y_j(t)_{gt}}  \right\|_1~~\textup{where}~~t\in [\Delta t : T_f]~~\textup{and}~~j\in [1, N]
\end{equation}
This metric is then used to compare the two methods. Based on~\cite{FORCASTBP} MAPE is not a suitable metric in the presence of outliers. To ensure a fair comparison we first, eliminate outlier results using the normal Z-score (or Six Sigma rules). This means that if the MAPE of a dataset performs significantly better or worse compared to others, it will not be included in the comparison of the two methods. Then we calculate the average of $\textup{MAPE}^{x_j}$ and $\textup{MAPE}^{y_j}$ over all test datasets for $x(t)$ and $y(t)$ respectively. In addition, we calculate the median and the 90th percentile. This will provide a more comprehensive understanding of the accuracy of the model. 
We conduct the same comparison without eliminating outlier results, using the sMAPE metric, which is more suitable in the presence of outliers (see Appendix ~\ref{appendix_1}).\\
During these tests, we observed that OASIS and Set Transformer may fail to predict the local dynamics, which can result in divergence. In some cases, this divergence is acceptable (Samples 11 and 21 in Fig.~\ref{fig:app_eg_extrapolation}), but we also found that it is possible for the model to completely fail to capture the dynamics, leading to infinite divergences. In this case, the dataset is not included in the comparison using the $\textup{MAPE}$ metric. To represent this, we added the divergence metric to highlight the model's ability to avoid complete divergences. \\
The results are presented in Tabs.~\ref{RESL_10_}, \ref{RESL_15_} and \ref{RESL_20_} which correspond to the 3 sub domains windows sizes as previously explained ($\Delta t_{10}$, $\Delta t_{15}$ and $\Delta t_{20}$), with four different extrapolation windows. This means that based on one sub domain window, we try to extrapolate for larger windows. We refer to these as extrapolation windows.\\

The primary attention lies in the divergence difference between the Set Transformer and OASIS frameworks, which increases with larger sub domain windows with 40-55\% of infinite divergences. In contrast, Set Transformer remains stable  with significantly lower divergence percentages. This phenomenon is understandable because the set encoding architecture attempts to encode all time series data to generate predictions. Additionally, as the quantity of data increases, so does the accuracy. We can also see that both models give very acceptable results for small extrapolation windows, this could be highly related to the training process of the two models, where we focus on predicting only the next sub domain dynamics. While the average of MAPE remains good for large extrapolation windows ($2\times\Delta t,~ 3\times\Delta t~~ \text{and}~ 4\times\Delta t$). The inaccuracy, or the 90th percentile represents the maximum MAPE error observed in 90\% of the test sets, increases significantly for OASIS, while it increases reasonably for Set Transformer maintaining acceptable extrapolation.\\

\begin{table}[h]
\centering\caption{Comparison between the two methods Set Transformer and OASIS in terms of MAPE error for extrapolation through local dynamics identification with a time window of 10 steps}
\label{RESL_10_}
\begin{tabular*}{\textwidth}{@{\extracolsep\fill}llrrrrrr}
\toprule%
& &\multicolumn{2}{@{}c@{}}{mean} & \multicolumn{2}{@{}c@{}}{90\%} & \multicolumn{2}{@{}c@{}}{divergence (\%)} \\\cmidrule{3-4}\cmidrule{5-6}\cmidrule{7-8}%
  &  & SETTR & OASIS & SETTR & OASIS & SETTR & OASIS \\
\midrule

1$\times \Delta t_{10}$ & $x$ & \textbf{0.43} & 1.01 & \textbf{0.76} & 3.75 & \textbf{5.36} & 44.64 \\
 & $y$ & \textbf{0.67} & 1.90 & \textbf{1.07} & 3.27 & \textbf{5.36} & 44.64 \\\midrule
2$\times \Delta t_{10}$ & $x$ & \textbf{1.35} & 2.35 & \textbf{3.63} & 9.73 & \textbf{5.36} & 42.86 \\
 & $y$ & \textbf{3.65} & 6.31 & \textbf{17.01} & 21.60 & \textbf{5.36} & 42.86 \\\midrule
3$\times \Delta t_{10}$ & $x$ & \textbf{1.93} & 3.21 & \textbf{4.21} & 10.28 & \textbf{7.14} & 41.07 \\
 & $y$ & \textbf{3.28} & 6.24 & \textbf{11.06} & 19.00 & \textbf{7.14} & 41.07 \\\midrule
4$\times \Delta t_{10}$ & $x$ & \textbf{4.09} & 5.44 & \textbf{12.20} & 15.40 & \textbf{3.57} & 39.29 \\
 & $y$ & \textbf{5.34} & 8.45 & \textbf{18.76} & 24.83 & \textbf{3.57} & 39.29 \\

\botrule
\end{tabular*}
\end{table}

\begin{table}[h]
\centering\caption{Comparison between the two methods Set Transformer and OASIS in terms of MAPE error for extrapolation through local dynamics identification with a time window of 15 steps}
\label{RESL_15_}
\begin{tabular*}{\textwidth}{@{\extracolsep\fill}llrrrrrr}
\toprule%
& &\multicolumn{2}{@{}c@{}}{mean} & \multicolumn{2}{@{}c@{}}{90\%} & \multicolumn{2}{@{}c@{}}{divergence (\%)} \\\cmidrule{3-4}\cmidrule{5-6}\cmidrule{7-8}%
  &  & SETTR & OASIS & SETTR & OASIS & SETTR & OASIS \\
\midrule

1$\times \Delta t_{15}$ & $x$ & 1.00 & \textbf{0.83} & 1.54 & \textbf{1.22} & \textbf{7.14} & 51.79 \\
 & $y$ & \textbf{0.88} & 2.92 & \textbf{1.50} & 4.63 & \textbf{7.14} & 51.79 \\\midrule
2$\times \Delta t_{15}$ & $x$ & \textbf{2.41} & 2.81 & \textbf{3.86} & 5.12 & \textbf{7.14} & 51.79 \\
 & $y$ & \textbf{2.26} & 5.49 & \textbf{2.75} & 8.98 & \textbf{7.14} & 51.79 \\\midrule
3$\times \Delta t_{15}$ & $x$ & \textbf{3.49} & 5.11 & \textbf{5.84} & 11.79 & \textbf{5.36} & 44.64 \\
 & $y$ & \textbf{2.88} & 7.83 & \textbf{5.56} & 19.44 & \textbf{5.36} & 44.64 \\\midrule
4$\times \Delta t_{15}$ & $x$ & \textbf{6.24} & 8.09 & \textbf{14.70} & 22.14 & \textbf{3.57} & 44.64 \\
 & $y$ & \textbf{5.20} & 11.80 & \textbf{16.38} & 38.38 & \textbf{3.57} & 44.64 \\

\botrule
\end{tabular*}
\end{table}

\begin{table}[h]
\centering\caption{Comparison between the two methods Set Transformer and OASIS in terms of MAPE error for extrapolation through local dynamics identification with a time window of 20 steps}
\label{RESL_20_}
\begin{tabular*}{\textwidth}{@{\extracolsep\fill}llrrrrrr}
\toprule%
& &\multicolumn{2}{@{}c@{}}{mean} & \multicolumn{2}{@{}c@{}}{90\%} & \multicolumn{2}{@{}c@{}}{divergence (\%)} \\\cmidrule{3-4}\cmidrule{5-6}\cmidrule{7-8}%
  &  & SETTR & OASIS & SETTR & OASIS & SETTR & OASIS \\
\midrule

1$\times \Delta t_{20}$ & $x$ & 2.42 & \textbf{1.09} & 3.99 & \textbf{1.84} & \textbf{7.14} & 55.36 \\
 & $y$ & \textbf{1.03} & 3.89 & \textbf{1.78} & 6.15 & \textbf{7.14} & 55.36 \\\midrule
2$\times \Delta t_{20}$ & $x$ & 5.10 & \textbf{4.41} & \textbf{8.79} & 9.15 & \textbf{7.14} & 50.00 \\
 & $y$ & \textbf{2.08} & 7.14 & \textbf{3.79} & 11.86 & \textbf{7.14} & 50.00 \\\midrule
3$\times \Delta t_{20}$ & $x$ & \textbf{9.90} & 10.17 & \textbf{15.92} & 23.57 & \textbf{7.14} & 44.64 \\
 & $y$ & \textbf{6.51} & 14.75 & \textbf{16.66} & 39.65 & \textbf{7.14} & 44.64 \\\midrule
4$\times \Delta t_{20}$ & $x$ & \textbf{9.28} & 10.25 & \textbf{14.16} & 24.59 & \textbf{3.57} & 53.57 \\
 & $y$ & \textbf{4.46} & 8.95 & \textbf{9.60} & 23.78 & \textbf{3.57} & 53.57 \\

\botrule
\end{tabular*}
\end{table}

\newpage

\begin{figure}[!htbp]
    \centering
    \includegraphics[width=0.9\linewidth]{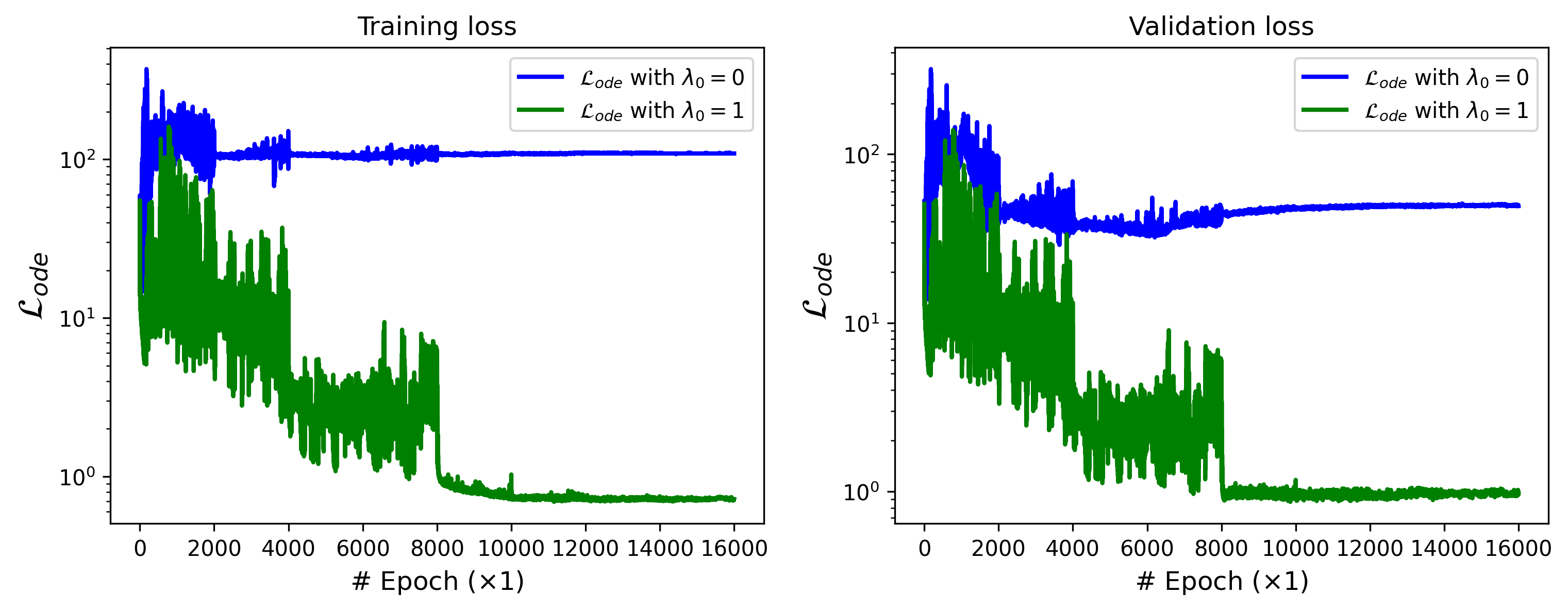}
    \caption{Loss comparison when using $\mathcal{L}_{\text{ode}}$ with $\lambda = 1$ versus when $\mathcal{L}_{\text{ode}}$ is not used, with $\lambda = 0$. For the Set Transformer method using the same architecture.}
    \label{lossode_train_valid}
\end{figure}

On the other hand, we concluded that adding the ODE loss $\mathcal{L}_{ode} = \left \| \frac{d\widetilde{X}}{dt} - \Theta(\widetilde{X},C)\hat{\Xi} \right \|_1$ as given in  Tab.~\ref{tab:trainprocess} significantly improves the results, especially in cases where the dynamics terms are unknown or only partially known, as well as in cases where the goal is to identify local dynamics valid for short time windows. The Fig.~\ref{lossode_train_valid} represents the ODE loss for training and validation using the Set Transformer method with the same architecture. When $\mathcal{L}_{ode}$ is used with $\lambda = 1$, and when $\lambda = 0$ indicates that the ODE loss is not used, we can easily observe the difference in convergence. This difference can be explained by the fact that $\mathcal{L}_{ode}$ assigns weighted attention to each coefficient. This means that if the contribution of a coefficient is significant, its loss $\mathcal{L}_{ode}$ will also be significant compared to other coefficients, and vice versa.

Based on this comparison, we can see the importance of using Set Encoding for local dynamics identification, which could be utilized for extrapolation, as demonstrated in the example above, or for other use cases such as predictive control. We demonstrated that OASIS may fail to capture the local dynamic, resulting in inaccurate results and divergence, while Set Transformer remains stable and leverages all available data to predict the local dynamics. These results could be improved by using variable sub domain windows during the training process, which would provide the Set Transformer method the ability to generate more accurate results and enhance extrapolation capabilities. This approach is demonstrated in the next example, where we aim to identify the global dynamics using an expanding time window. 
Furthermore, using this approach with noisy data is going to be challenging, which is related to the high sensitivity of SINDy to noise~\cite{SINDYNOISE}. Additionally, in this case, we only aim to identify a local dynamic, which means that the identification process is more challenging. Moreover, in extrapolation use cases, it is essential to update the model frequently to obtain accurate results. Using the predictions of Set Transformer (or OASIS) and then an ODE solver to extrapolate poses two challenges. The first one is related to the training process, which involves the generation of training data and using SINDy for the identification of local dynamics with noisy data, which is challenging. Furthermore, even slight noise in the initial conditions may lead to inaccurate extrapolation or complete divergence.

\subsection{Application 2 : Global model identifications (Deep Set vs Set Transformer)}\label{application2}

\subsubsection{Data generation}
We generate a dataset of 1024 Lorenz systems with variables parameters  $\left [ \sigma,~\rho,~\beta \right ]$ , and  initial conditions $\left [ x_0,~y_0,~z_0 \right ]$ as described in Tab.~\ref{tab:res2DataSet} using LSODA~\cite{LSODA} methods implemented in SciPy Python package. The parameters and initial conditions are sampled using the Sobol~\cite{SOBOL, LSODA} method implemented in the SciPy Python package. As the number of possible combinations of parameters and initials conditions is large and taking into consideration the chaotic behavior of the Lorenz system, we use 70\% data to train the model and the remaining 30\% for validation and testing. \\

\begin{figure}[!htbp]
    \centering
    \includegraphics[width=1.1\linewidth]{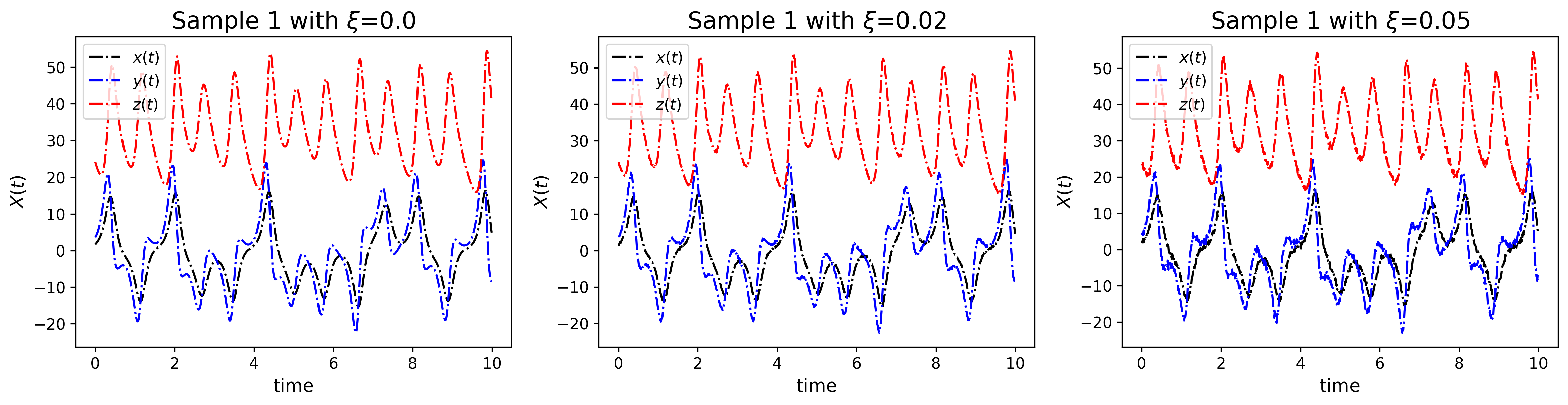}
    \caption{Example of the Lorenz system with different noise levels, left 0\% ($\xi=0$), middle 2\% ($\xi=0.02$), and right 5\% ($\xi=0.05$)}
    \label{lorenz3lvlsnoise}
\end{figure}

\begin{table}[!htbp]
\centering
\caption{Range of Variables}\label{tab:res2DataSet}
\begin{tabular}{@{}ll@{}}
\toprule
\textbf{Variables} & \textbf{Range} \\
\hline
\(\sigma\) variations & [8; 12] \\

\(\rho\) variations & [20; 35] \\

\(\beta\) variations & [0; 4] \\

\(x_0\) variations & [-5; 5] \\

\(y_0\) variations & [4; 50] \\

\(z_0\) variations & [5; 15] \\

\(t\) train and validation & [0, 10[ \\

\(t\) test & [0, 20[ \\
\botrule
\end{tabular}
\end{table}

To study the robustness of the models concerning data quality, we generate 3 datasets with three levels of Gaussian noise. $\xi\in\left [0, 0.02, 0.05 \right ]$ as described in Eq.~\ref{eq:eqnoise} :
\begin{equation}\label{eq:eqnoise}
\left\{\begin{matrix}
x(t) &=& x(t) + \xi\epsilon_x ~~where~~\epsilon_x \sim \mathcal{N}(0,\,\sigma^{2}_x)\,\\
y(t) &=& y(t) + \xi\epsilon_y ~~where~~\epsilon_y \sim \mathcal{N}(0,\,\sigma^{2}_y)\,\\
z(t) &=& z(t) + \xi\epsilon_z ~~where~~\epsilon_z \sim \mathcal{N}(0,\,\sigma^{2}_z)\,
\end{matrix}\right.
\end{equation}

Fig.~\ref{lorenz3lvlsnoise} shows an example with three levels of noise. Using the same parameters and initial conditions, we then compute the approximation of the derivative with respect to time $\widetilde{\frac{\mathrm{d}X }{\mathrm{d} t}}~=~\left [\widetilde{\frac{\mathrm{d}x }{\mathrm{d} t}},~\widetilde{\frac{\mathrm{d}y }{\mathrm{d} t}},~\widetilde{\frac{\mathrm{d}z }{\mathrm{d} t}} \right ]$ using total variation method~\cite{chartrand2011numerical} for noisy data and central difference (Eq.~\ref{eq:ctrldiff}) for clean data. \\

\begin{equation}\label{eq:ctrldiff}
f'(x)\simeq {\frac {f(x+h)-f(x-h)}{2h}}
\end{equation}

\begin{figure}[!htbp]
    \centering
    \includegraphics[width=1.\linewidth]{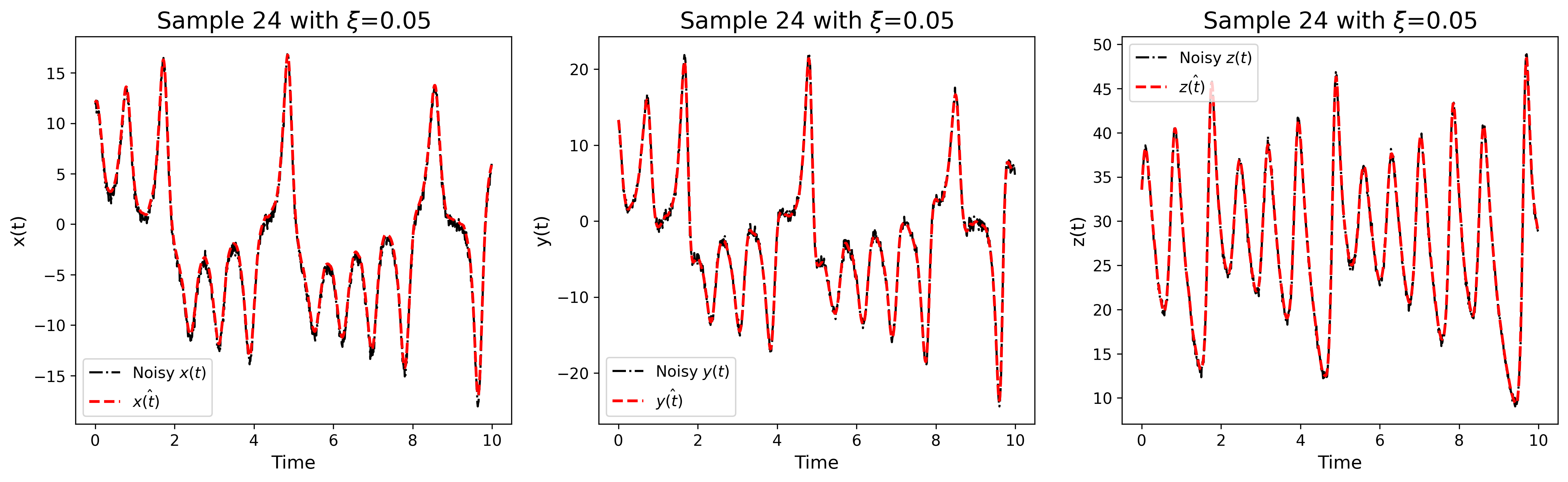}
    \caption{Example of the Lorenz system: comparison of noisy data and denoised data using the Total Variation method}
    \label{application2denoise}
\end{figure}

Afterwards, we compute $\widetilde{X}~=~\left [\widetilde{x},~\widetilde{y},~\widetilde{x} \right ]$, which represents the approximation of the integral of $\widetilde{\frac{\mathrm{\partial}X }{\mathrm{\partial} t}}$ using the cumulative trapezoidal integration method (Eq.~\ref{eq:trapez}) (with the SciPy Python package). 
\begin{equation}\label{eq:trapez}
\widetilde{X}~=~\int_{a}^{b}\widetilde{\frac{\mathrm{\partial}X }{\mathrm{\partial} t}}dt~\simeq ~\frac{1}{2}\sum_{n=1}^{N-1}(t_{n+1}-t_n)[\widetilde{\frac{\mathrm{\partial}X }{\mathrm{\partial} t}}_{n+1}+\widetilde{\frac{\mathrm{\partial}X }{\mathrm{\partial} t}}_n]
\end{equation}
This approach contributes to noise reduction in the data (Fig.~\ref{application2denoise}), it will also enable the models to learn effective denoising techniques (further explanations are given in  section~\ref{application2Trainprocess}). PySINDy (Python package) is then used to compute the approximation of the parameters for each dataset. In this application we assume that the ODE search library $\Theta$ is known and well defined, our main focus is on calculating the parameters $\left [ \sigma_k,~\rho_k,~\beta_k \right ]$. The identification process is defined as in Eq.~\ref{eq:sindy_app2} .

\begin{equation}\label{eq:sindy_app2}
\left[\begin{matrix}\widetilde{\frac{\mathrm{d}x }{\mathrm{d} t}}\\
\widetilde{\frac{\mathrm{d}y }{\mathrm{d} t}}+\widetilde{y} +\widetilde{x}\widetilde{z} \\
\widetilde{\frac{\mathrm{d}z }{\mathrm{d} t}}-\widetilde{x}\widetilde{y}
\end{matrix}
\right]~=~  \left[
\widetilde{x}-\widetilde{y},~\widetilde{y},~\widetilde{z}\right]~\widetilde{\Xi}~=~ \Theta(\widetilde{x},~\widetilde{y},\widetilde{z})~\widetilde{\Xi}
\end{equation}

\subsubsection{Train process and criteria of comparison}\label{application2Trainprocess}
We implement a consistent training regime for both models (Deeps Set and Set Transformer), employing varied architectures for both methods. Following this, we conduct a comparison of the optimal architecture configurations for each method. A core principle underlying the utilization of Set Encoding models is the integration of adaptability, enabling predictions that dynamically adjust to an increasing volume of data over time. We employ a sliding window approach with increasing widths. Through this approach, the models are designed to adaptively incorporate new information for updating predictions of Lorenz parameters.\\

\begin{table}[!htbp]
\centering

\caption{Training process summary}\label{tab:trainprocess2}
\begin{tabular}{@{}lc|c@{}}
\toprule
                      & \multicolumn{1}{c}{\textbf{Deep Set}} & \textbf{Set Transformer} \\ \hline
\textbf{Optimizer}        & \multicolumn{2}{c}{Adam}              \\ \midrule
\textbf{Loss function}           & \multicolumn{2}{c}{$\left \| \widetilde{\Xi} -\hat{\Xi} \right \|_2 + \lambda_0\left \| \frac{d\widetilde{X}}{dt} - \Theta(\widetilde{X},C)\hat{\Xi} \right \|_1$}              \\ \midrule
\textbf{Learning rate}           & \multicolumn{2}{c}{$\left [ 4\times10^{-4},~5\times10^{-5},~10^{-5},~5\times10^{-6},~10^{-6} \right ]$}              \\ \midrule
\textbf{Epochs}           & \multicolumn{2}{c}{[400,~400,~400,~400,~400]}              \\ \midrule
\textbf{Input Shape}  & \multicolumn{2}{c}{$\left [Batchsize,~time,~Features \right ] = [Batchsize,~variable,~4]$}         \\ \midrule
\textbf{Output Shape} & \multicolumn{2}{c}{$\left [Batchsize,~3 \right ]$}                            \\ \botrule
\end{tabular}
\end{table}

The model inputs are $[t,~X_k(t)]~=~[t,~x_k(t),~y_k(t),~z_k(t)]$, which represent the noisy data, we added the time dimension, which will be encoded during the pooling phase. This will enable the model to better comprehend the dynamic while increasing input size, and hold good accuracy with variables times steps $\Delta t$. The output is defined as $\left [ \sigma_k,~\rho_k,~\beta_k \right ]$. The training process is summarized in Tab.~\ref{tab:trainprocess2}. We employ the Adam~\cite{ADAM} optimizer for training both models, with a step decay learning rate. The loss function comprises two components. The first part represents the loss associated with parameters loss  where $\widetilde{\Xi}$ and $\hat{\Xi}$ denote the fitted parameters using SINDy (Eq.~\ref{eq:sindy_app2}) based on denoised data and the predicted parameters. Note here that $\widetilde{\Xi}$ may differ from $\Xi$, the true parameters vector. The second part represents the loss associated with the ODE loss utilizing denoised variables, in this study, we fix $\lambda_0$ to 1. Training both methods to predict denoised parameters enables the model to additionally learn denoising techniques while processing noisy input data. Both methods are trained and validated on the same dataset. The best model will be selected based on its accuracy in validation; subsequently, the best model instances are chosen for a comparative analysis between the two models, for more details see appendix~\ref{appendix_3}.\\
The comparison of instances of the two models will be based on their ability to deliver accurate and stable predictions, while considering the variability in both the quality and quantity of data. Note here that we trained individual instances to predict each parameter of the Lorenz systems, which helps the models converge faster.

\subsubsection{Comparison and discussion}
The instances comparison are based on three features : \\
\begin{enumerate}
\item \textbf{Accuracy} : the accuracy of selected instances is assessed using $R^2$-score metric~\cite{chicco2021coefficient} defined as : 
\begin{equation}\label{eq:r2score}
R^2 = 1 - \frac{\sum _i (y_i-\hat{y}_i)^2}{\sum _i (y_i-\bar{y}_i)^2} ,
\end{equation}
which measures how well the predicted values match the actual values. In our context, it quantifies the precision of predicting Lorenz parameters.  The closer the $R^2$ score is to 1, the better the accuracy. It is important to note that our comparison involves evaluating the model predictions against the true Lorenz parameters, and not the fitted SINDy parameters $\hat{\Xi}$. We assess the instances' capability to provide accurate predictions with variable input sizes, we employ the weighted average of $R^2_{w.a}$ defined  as : 
\begin{equation}\label{eq:war2}
R^2_{w.a} = \sum _i\frac{1/n_i}{F_T}R^2_{n_i} ~~~\textup{where}~~ F_T=\sum_i 1/n_i ,
\end{equation}
which assigns more importance to predictions with fewer data points, where $n_i \in [100,~300,~500,~700,~900]$. This study is based on an expanding window approach, with two windows utilized: one for interpolation study where $t \in [0,~10]$, and another for extrapolation where $t \in [10,~20]$, which is never used during training or validation.

Results of $R^2_{w.a}$ are presented in Fig.~\ref{R2APP2}. The overall performance of both methods is good for both interpolation and extrapolation ($R^2_{w.a} > 0.8$). The accuracy drops for extrapolation, which is related to new Lorenz patterns unseen during the training process. It is also related to the time dimension used during training, limited to $t \in [0,10]$. In interpolation, Method Deep Set yields slightly better results, while in extrapolation, Method Set Transformer tends to be more stable (results do not change much). An exception is the prediction of $\sigma$, for which the accuracy drops for both methods. This can be explained by the data generation process (Eq.~\ref{eq:sindy_app2}), where we can observe that the prediction of $ \sigma $ is based on $ x - y $, whereas for the prediction of $ \beta $ and $ \rho $, it is based directly on $ y $ and $ z $, respectively, which are the input of the models, making it easier to learn. Furthermore, in extrapolation, Set Transformer performs better in the prediction of $\sigma$, while in interpolation, Deep Set was slightly better. 

We can also observe that for both interpolation and extrapolation, both methods demonstrate very good robustness to different noise levels. We present the $R^2_{n_i}$ with respect to the quantity of data in appendix~\ref{appendix_2} (Fig.~\ref{APP2SIGMA},~\ref{APP2BETA} and \ref{APP2RHO}).

\begin{figure}[!htbp]
    \centering
    \includegraphics[width=1.\linewidth]{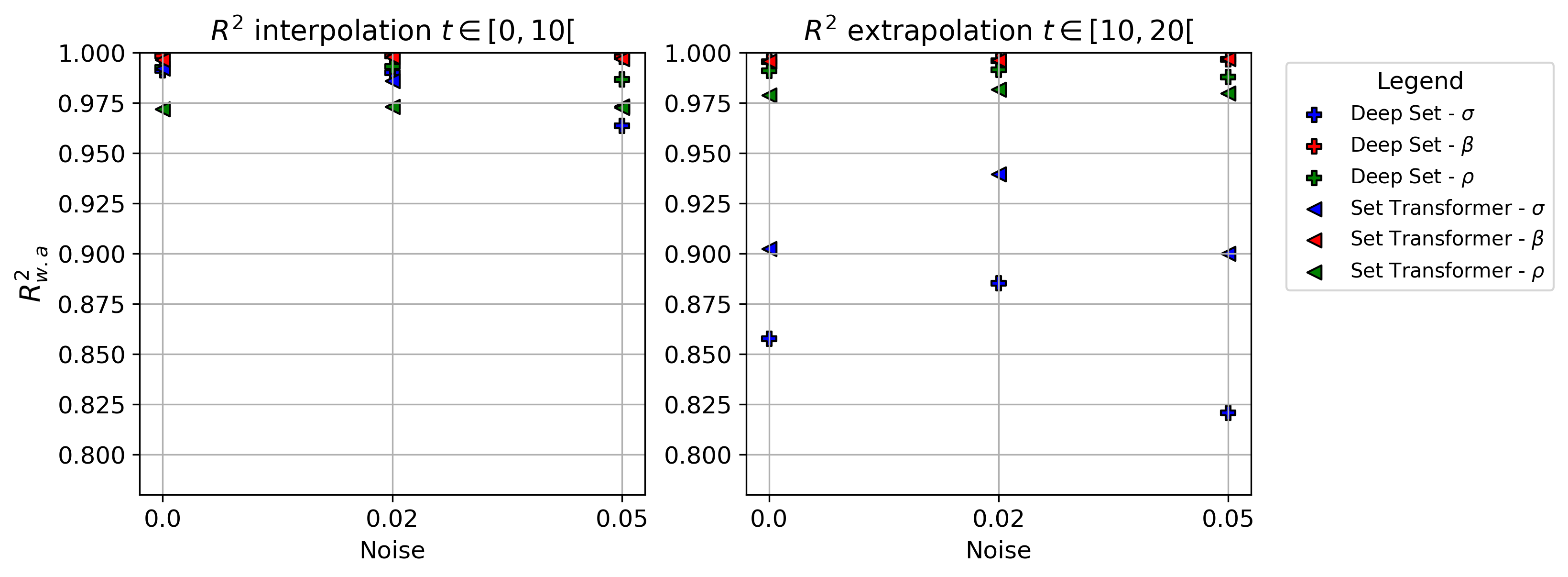}
    \caption{Accuracy results using $R^2_{w.a}$ metric for Deep Set and Set Transformer instances for $\sigma$, $\beta$ and $\rho$. On the left interpolation results for $t \in [0,10[$, and on the right extrapolation results for $t \in [10,20[$.}
    \label{R2APP2}
\end{figure}

\item \textbf{Inference time} : Both trained models could be used directly for online applications. This is because the inference time is in the order of a few milliseconds on AuthenticAMD AMD EPYC 9474F@3.6Ghz 48-Core Processor, making these models more suitable for applications like the extrapolation of state variables~\cite{OASISP} or online control~\cite{OASIS}. These methods also can dynamically use new data to enhance the accuracy of the prediction or to reevaluate the prediction in case of changes in the system dynamics. 
\item \textbf{Training} : Based on the comparison of both methods, we observed that the implementation of Deep Set is easier compared to Set Transformer. Deep Set relies on simple dense layers, resulting in fewer hyperparameters, thus facilitating hyperparameter tuning. More details about the instance selection are provided in Appendix \ref{appendix_3}.
\end{enumerate}

\textcolor{black}{
\subsection{Application 3: Characterizing Abnormalities in 1D Heat-Transfer Using Deep Set Model}\label{App3}
\subsubsection{Data generation}
For this application we aim to train a Deep Set model to approximate online the diffusivity value $\alpha$ from temperature data, we also challenge the model to characterize a local decrease of diffusivity. We start by generating a dataset using finite difference methods (implicit Euler method) with different diffusivity values $\alpha$. The 1D heating equation is defined as in Eq.~\ref{eq:pde1d} in this case $\alpha (z)$ is space-dependent. The initial temperature value is $T(t=0, z) = 20~^\circ C$, $L = 0.01~m$, and the heat flux is defined as function as follow (Eq.~\ref{eq:heating_flux}) :
\begin{equation}\label{eq:heating_flux}
\left\{\begin{matrix}
-\frac{\partial  T}{\partial z}|_{z=0} &=& \frac{q_0}{k} = 30000~K/m~~~&\textbf{if}&~~T(z=0)<T_{max}\\
-\frac{\partial  T}{\partial z}|_{z=0} &=& 0~K/m~~~&\textbf{if}&~~T(z=0)>T_{max}\\
\end{matrix}\right.
\end{equation}
Where $T_{max}=500~^\circ C$, which simply means that we impose a regularization to stop the heating if $T>T_{max}$. The $\alpha (z)$ is defined as in Fig.~\ref{EXMPDEFCAR}, this function could be simplified as three characteristics, $\alpha_{ref}$ which is the normal diffusivity value, $\mathrm{G} = \mathrm{center_{abnormality}}$ which presents the center of abnormality and $\mathrm{ratio} = \frac{\alpha_{ref}}{\alpha_{abnormality}}$ which is simply the ratio between normal and abnormal diffusivity.
The $\alpha (z)$ function defined as (Eq.~\ref{eq:alpha_z}), where $A$ and $B$ are two values that help define $\alpha_{abnormality}$:
\begin{equation}\label{eq:alpha_z}
\alpha(z) = \alpha_{ref} - Ae^{-B(\mathrm{G}-z)^6}
\end{equation}
\begin{figure}[!htbp]
    \centering
    \includegraphics[width=0.6\linewidth]{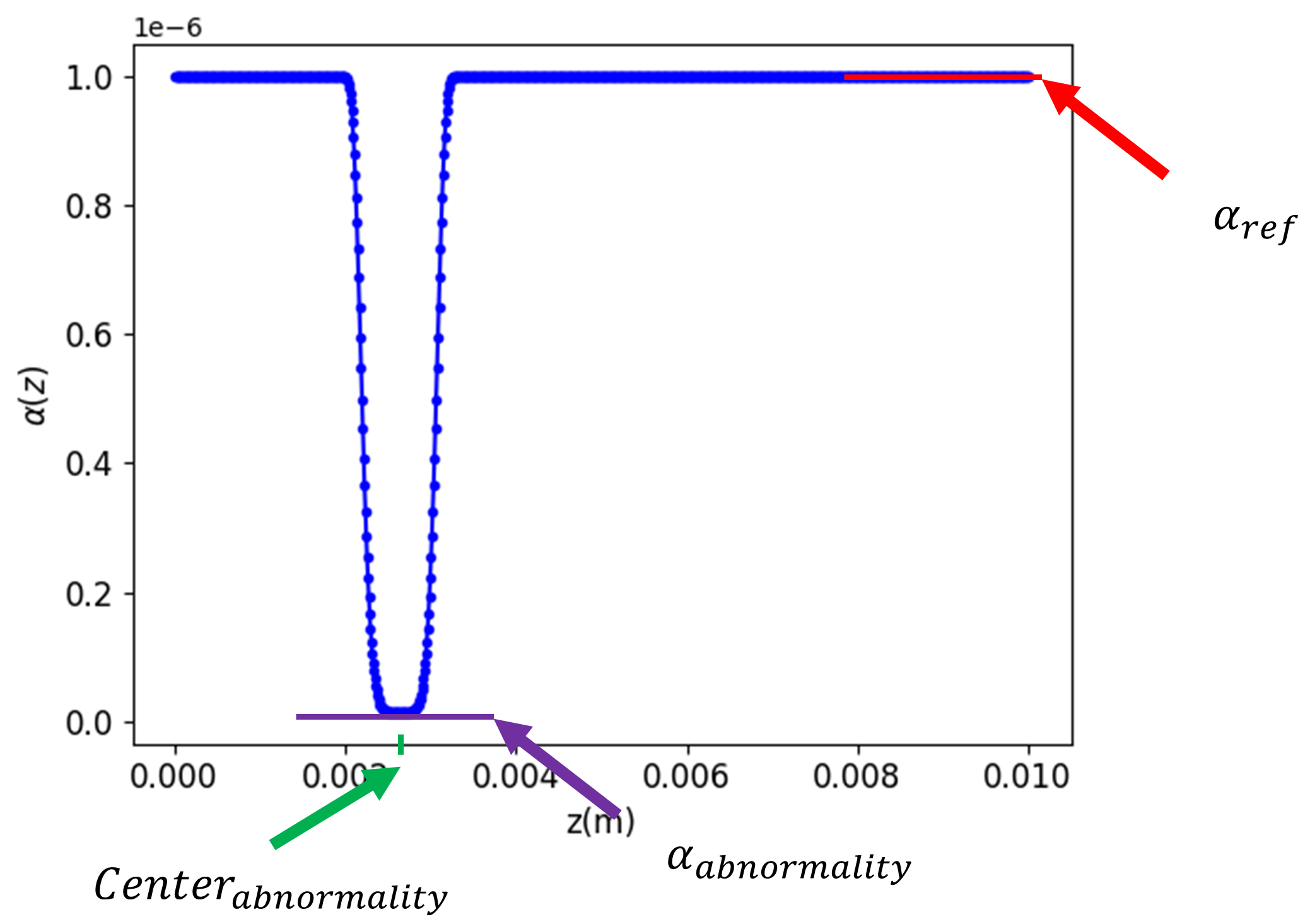}
    \caption{Example of the function diffusivity $\alpha(z)$ with respect to space $z$, which allows the characterization of abnormalities.}
    \label{EXMPDEFCAR}
\end{figure}
We generate $N$ datasets with different combinations of the three characteristics defined previously, where $N=300$, where $70\%$ of this dataset is used for training and $30\%$ for validation, we also generate separately a $N_{test}=150$ test datasets to evaluate the performance of the model, the distribution of training, validation and test characteristics is defined as follows (Tab.~\ref{tab:rngcharacteristics}) :
\begin{table}[!htbp]
\centering
\caption{Range of characteristics for train, validation and test}\label{tab:rngcharacteristics}
\begin{tabular}{@{}lll@{}}
\toprule
\textbf{Variables} & \textbf{Range} & \textbf{Distribution}\\
\hline\\
$\alpha_{ref}$  & [$8\times10^{-7};~1.2\times10^{-6}$] & $\mathcal{N}(10^{-6},~~(5\times10^{-8})^2)$\\
$\mathrm{ratio}$  & [$40;~80$] & $\mathcal{N}(60,~~(8)^2)$\\
$\mathrm{G}$  & [$0.001;~0.009$] & Sobol\\
Time $t$  & [$0;~200$] & - \\
\botrule
\end{tabular}
\end{table}
\subsubsection{Training process}
Using these datasets, we train a Deep Set model, where the input
$X_i$ combines a time series of temperature at two positions $z=0$ and $z=L=0.01~m$ and the corresponding time $t$ (Example in Fig.~\ref{INPUTEXMP}), where $\Delta t = 1~s$. In this study we limited the model input temperature information to only two positions, the input $X_{i}$ which corresponds to the $i^{th}$ dataset is defined as (Eq.~\ref{eq:inputdata}) :
\begin{equation}\label{eq:inputdata}
X_i~=~\left [z_i,~t_i,~T_i \right ]~=~\begin{bmatrix}
z_{i,1} & t_{i,1} & T(t_{i,1},~z_{i,1})\\ 
z_{i,1} & t_{i,2} & T(t_{i,2},~z_{i,1})\\ 
&\cdots &\\ 
&\cdots& \\ 
z_{i,k} & t_{i,j} & T(t_{i,j},~z_{i,k})\\ 
z_{i,k} & t_{i,j+1} & T(t_{i,j+1},~z_{i,k})\\ 
&\cdots &\\
&\cdots &\\  
z_{i,L} & t_{i,T-1} & T(t_{i,T-1},~z_{i,L})\\
z_{i,L} & t_{i,T} & T(t_{i,T},~z_{i,L}) 
\end{bmatrix} ~~ \text{where} ~~ k\in\{1, L\},~~j\in[1, T]
\end{equation}
Thus the model input and output could be defined as $F(X_i;~~\Theta) = [\text{center}_i, \text{ratio}_i, \alpha _i]$, where $\Theta$ are the trainable parameters of Deep Set model. The details about the used architecture are given in~\ref{appendix_last}. We used a time-expanding window, which gives the model a flexible approximation of the characteristics using available information, which is improving as new data is available. The loss function is defined as (Eq.~\ref{eq:trainloss}):
\begin{equation}\label{eq:trainloss}
\left\{\begin{matrix}
\mathscr{L}_{Total} &=& \mathscr{L}_{\text{center}}+ \mathscr{L}_{\text{ratio}} + \mathscr{L}_{\alpha_{ref}} + \lambda\mathscr{L}_{\Theta}\\
\mathscr{L}_{\text{center}} &=& \left\| \mathrm{G}_{true} - \mathrm{G}_{prediction}\right\|_2\\
\mathscr{L}_{\text{ratio}} &=& \left\| \mathrm{ratio}_{true} - \mathrm{ratio}_{prediction}\right\|_2\\
\mathscr{L}_{\alpha_{ref}} &=& \left\| \alpha _{true} - \alpha _{prediction}\right\|_2\\
\mathscr{L}_{\Theta} &=& \left\|\Theta\right\|_1\\
\end{matrix}\right.
\end{equation}
, which is composed of four losses:  $\mathscr{L}_{\text{center}}$, $\mathscr{L}_{\text{ratio}}$,  $\mathscr{L}_{\alpha}$ and $\mathscr{L}_{\Theta}$ , which correspond to the losses of each characteristic and the regularization of the model which helps avoid overfitting problems, in this application we found that $\lambda = 0.01$ gives better generalization.\\
\begin{figure}[!htbp]
    \centering
    \includegraphics[width=0.6\linewidth]{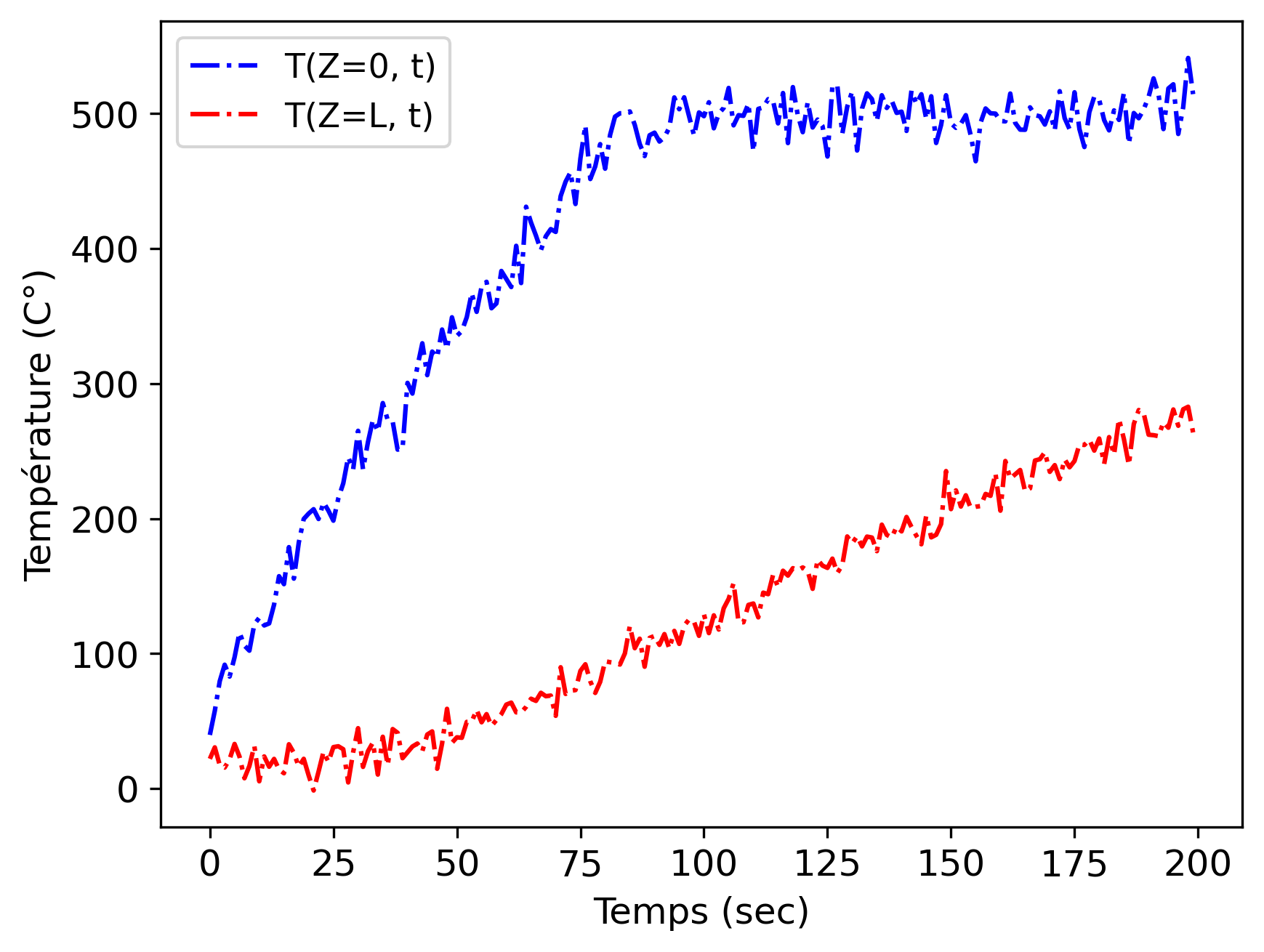}
    \caption{Example of the input temperature with noise of the Deep Set model.}
    \label{INPUTEXMP}
\end{figure}
We also study the model robustness to noise, for that we tested the model for four noise levels defined as (Eq.\ref{eq:app3eqnoise}) where $\xi \in [0, 0.02, 0.05, 0.1]$ and $i \in [0,~N]$:
\begin{equation}\label{eq:app3eqnoise}
T_{i}(t) = T_{i}(t) + \xi\epsilon_T \quad \text{where} \quad \epsilon_T \sim \mathcal{N}(0,\,\sigma^{2}_{T_{i}})
\end{equation}
\subsubsection{Results and discussion}
We trained different models for the four different noise levels $\xi$, we then select the best model based on the loss of the validation datasets, the selected models are then tested using the test datasets. The selected model architecture is given in appendix~\ref{appendix_last}. We assess the model accuracy with respect to the quantity data and robustness to noise using the $R^2$ metrics (Eq.~\ref{eq:r2score}), we also use the same temperature position used for the training and validation, the results are :
\begin{itemize}
\item \textbf{Accuracy and robustness} : We use the same accuracy assessment $R^2_{w.a}$ as in application~\ref{application2} :
\begin{equation}\label{eq:war2}
R^2_{w.a} = \sum _i\frac{1/n_i}{F_T}R^2_{n_i} ~~~\textup{where}~~ F_T=\sum_i 1/n_i ,
\end{equation}
, for this application we test the models for different data quantities where $t \in [0,~200]$ and data quantities $n_i \in [50,~100,~150,~200]$.\\
\begin{figure}[!htbp]
    \centering
    \includegraphics[width=0.65\linewidth]{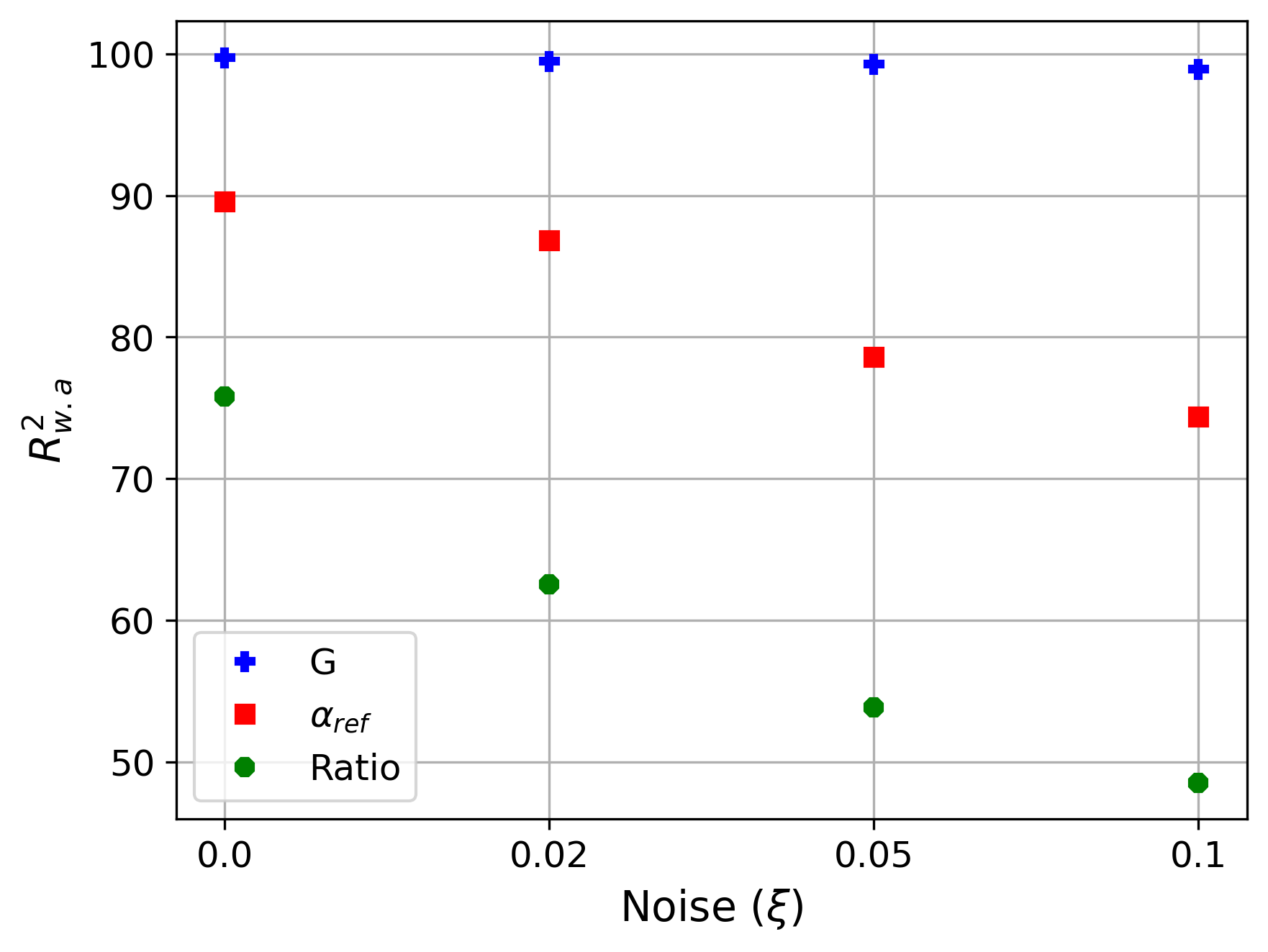}
    \caption{Accuracy results using $R^2_{w.a}$ metric for Deep Set for abnormalities characterization : $\mathrm{G}$, $\alpha_{ref}$ and $\mathrm{ratio}$, for 4 different noise levels from $\xi = 0.0$ to $\xi = 0.1$}
    \label{APP3AWRESL}
\end{figure}
From these results (Fig.~\ref{APP3AWRESL}, and.~\ref{APP3ALLRESL}), we can see that the model can easily predict the $\mathrm{G}$ of abnormality and remains stable with respect to noise, also the accuracy of the prediction of $\alpha _{ref}$ is good and decreases reasonably with the increase of noise, we believe that this is related to confusion related to the added noise which makes the precision of diffusivity very difficult, this is also related to the fact that the range of diffusivity is small while the noise levels represent a higher $\Delta T_{noise}$ related to noise (for $\xi=0.1$ $\Delta T_{noise} \in [-40, 40]$) which confuse the training of the model. Finally, the prediction of the $\mathrm{ratio}$ is acceptable for low noise levels but decreases as the noise level increases. We believe that this is related to the lack of information (only two temperature positions) and the confusion caused by the noise level. We also believe that increasing the number $N$ of datasets will improve the model accuracy. More details on the accuracy of the model with respect to data quantity are given in appendix~\ref{appendix_4}.
\item \textbf{Discussion} : In this application we focused on the Deep Set architecture, nevertheless Set Transformer, which gives very good results for Lorenz chaotic systems, could also be applied to this application and is expected to give good results. Also, in this application, we did not add a physics loss $\mathscr{L}_{PDE}$ simply because the approximation of $\frac{\partial (\alpha(z) \partial T)}{\partial z^2}$ in the PDE (Eq.~\ref{eq:pde1d}) is not an easy task. Specifically, in this case where we use only two separated positions, the approximation of the second derivative with respect to space is challenging. This results in less robustness to noise compared to the second application. This problem will be addressed in a subsequent study where we will attempt to improve this method for application to PDE problems.
\end{itemize}}

\newpage

\section{Conclusion and perspectives}\label{SEC6}
In this paper, we presented a combination of two architectures of Set Encoding with the methods of SINDy and SINDYc. The Set Encoding methods are initially trained on a dataset generated by SINDy (or SINDYc) and subsequently employed in real-time to swiftly and accurately predict system dynamics parameters, which could be used for applications such as extrapolating state variables or system control. Our primary contribution lies in encoding variable-sized data sequences using an appropriate architecture, thereby enabling the model to better utilize all temporal information to enhance precision. Additionally, we propose a more suitable loss function. In application 1, we illustrate the need to incorporate the $\mathcal{L}_{ode}$ component in the loss function, which enhances the model training process. Our approach can be applied to local identification and global dynamics, as demonstrated in application 2 \textcolor{black}{and 3}. In the second application, we suggest a training approach on noisy data, demonstrating that both methods yield robust and stable results with different noise levels. However, this approach relies on datasets generated by SINDy, implying that accuracy is directly dependent on the precision of SINDy (or SINDYc).  \textcolor{black}{In the third application, we addressed a PDE example and showed that the model gives interesting results. Nevertheless, the generation of the training datasets is more challenging, especially when dealing with limited and noisy datasets in the context of partial differential equation (PDE) systems, where calculating derivatives in the presence of noise is particularly challenging.}. To address this issue, an approach akin to PINN-SR~\cite{PINNSR} can be employed, extending our methodology to handle ODEs and PDEs with noisy and limited data quantities. In addition, the current Set Encoding architecture cannot handle cases where terms in differential equations are prone to deletion or addition in real-time applications. This implies that the models cannot activate or deactivate terms. To address this challenge, one can integrate a new decoding block with an appropriate activation function (such as a sigmoid function), which would be responsible for activating and deactivating dynamic system terms. Finally, we believe that a comparison between Deep Set and Set Transformer in online nonlinear dynamics identification is necessary to better understand the difference between these two methods. While Deep Set encodes data using classical neural network architecture, Set Transformer employs attention mechanisms, which enable better learning of interactions among input data. This has shown interesting results in many application fields. However, these advantages come with drawbacks such as increased training time and architectural complexity.

\appendix

\newcommand{\abs}[1]{\lvert#1\rvert}
\section{Appendix}

\subsection{Using sMAPE metrics for comparison between OASIS and Set Transformer}\label{appendix_1}
We present the results based on the sMAPE metric (Eqs.~\ref{eq:OA_sMAPE_error_x},~and~\ref{eq:OA_sMAPE_error_y}). This metric is more suitable for evaluating the forecasting of time series with outliers. In this case, it is not necessary to eliminate outliers; we follow the same approach as that presented in section~\ref{app1compar}. According to the results presented in Tabs.~\ref{RESL_10_smape}, \ref{RESL_15_smape} and \ref{RESL_20_smape}, we can see a close resemblance to the results presented using MAPE with outlier elimination.\\\\

\begin{equation}
\label{eq:OA_sMAPE_error_x}
\textbf{sMAPE}_{Total}^{x_j} =  \left\| 200\times\frac{x_j(t)_{p} - x_j(t)_{gt}}{\abs{x_j(t)_{gt}} + \abs{x_j(t)_{p}}}  \right\|_1~~\textup{where}~~t\in [\Delta t : T_f]~~\textup{and}~~j\in [1, N]
\end{equation}

\begin{equation}
\label{eq:OA_sMAPE_error_y}
\textbf{sMAPE}_{Total}^{y_j} =  \left\| 200\times\frac{y_j(t)_{p} - y_j(t)_{gt}}{\abs{y_j(t)_{gt}} + \abs{y_j(t)_{p}}}  \right\|_1~~\textup{where}~~t\in [\Delta t : T_f]~~\textup{and}~~j\in [1, N]
\end{equation}

\begin{table}[h]
\centering\caption{Comparison between the two methods Set-Transformer and OASIS in terms of sMAPE error for extrapolation through local dynamics identification with a time window of 10 steps}
\label{RESL_10_smape}
\begin{tabular*}{\textwidth}{@{\extracolsep\fill}llrrrrrr}
\toprule%
& &\multicolumn{2}{@{}c@{}}{mean} & \multicolumn{2}{@{}c@{}}{90\%} & \multicolumn{2}{@{}c@{}}{divergence (\%)} \\\cmidrule{3-4}\cmidrule{5-6}\cmidrule{7-8}%
  &  & SETTR & OASIS & SETTR & OASIS & SETTR & OASIS \\
\midrule

1$\times \Delta t_{10}$ & $x$ & \textbf{0.47} & 1.23 & \textbf{1.11} & 3.58 & \textbf{5.36} & 44.64 \\
 & $y$ & \textbf{1.46} & 3.61 & \textbf{1.23} & 6.06 & \textbf{5.36} & 44.64 \\\midrule
2$\times \Delta t_{10}$ & $x$ & \textbf{1.96} & 3.40 & \textbf{6.07} & 12.20 & \textbf{5.36} & 42.86 \\
 & $y$ & \textbf{4.45} & 6.44 & \textbf{18.69} & 19.46 & \textbf{5.36} & 42.86 \\\midrule
3$\times \Delta t_{10}$ & $x$ & \textbf{2.31} & 4.23 & \textbf{5.78} & 13.59 & \textbf{7.14} & 41.07 \\
 & $y$ & \textbf{4.16} & 6.23 & \textbf{14.40} & 18.15 & \textbf{7.14} & 41.07 \\\midrule
4$\times \Delta t_{10}$ & $x$ & \textbf{4.51} & 5.87 & \textbf{13.44} & 16.14 & \textbf{3.57} & 39.29 \\
 & $y$ & \textbf{6.85} & 9.07 & 24.34 & \textbf{21.02} & \textbf{3.57} & 39.29 \\

\botrule
\end{tabular*}
\end{table}

\begin{table}[h]
\centering\caption{Comparison between the two methods Set-Transformer and OASIS in terms of sMAPE error for extrapolation through local dynamics identification with a time window of 15 steps}
\label{RESL_15_smape}
\begin{tabular*}{\textwidth}{@{\extracolsep\fill}llrrrrrr}
\toprule%
& &\multicolumn{2}{@{}c@{}}{mean} & \multicolumn{2}{@{}c@{}}{90\%} & \multicolumn{2}{@{}c@{}}{divergence (\%)} \\\cmidrule{3-4}\cmidrule{5-6}\cmidrule{7-8}%
  &  & SETTR & OASIS & SETTR & OASIS & SETTR & OASIS \\
\midrule

1$\times \Delta t_{15}$ & $x$ & \textbf{1.14} & 1.76 & \textbf{1.87} & 3.01 & \textbf{7.14} & 51.79 \\
 & $y$ & \textbf{1.84} & 4.62 & \textbf{1.81} & 8.41 & \textbf{7.14} & 51.79 \\\midrule
2$\times \Delta t_{15}$ & $x$ & \textbf{2.69} & 3.78 & \textbf{5.07} & 10.55 & \textbf{7.14} & 51.79 \\
 & $y$ & \textbf{3.37} & 5.76 & \textbf{7.32} & 12.71 & \textbf{7.14} & 51.79 \\\midrule
3$\times \Delta t_{15}$ & $x$ & \textbf{4.60} & 6.13 & \textbf{6.03} & 13.73 & \textbf{5.36} & 44.64 \\
 & $y$ & \textbf{3.71} & 8.48 & \textbf{9.53} & 20.58 & \textbf{5.36} & 44.64 \\\midrule
4$\times \Delta t_{15}$ & $x$ & \textbf{7.40} & 9.56 & \textbf{14.88} & 24.25 & \textbf{3.57} & 44.64 \\
 & $y$ & \textbf{7.05} & 10.83 & \textbf{18.18} & 31.84 & \textbf{3.57} & 44.64 \\

\botrule
\end{tabular*}
\end{table}

\begin{table}[h]
\centering\caption{Comparison between the two methods Set-Transformer and OASIS in terms of sMAPE error for extrapolation through local dynamics identification with a time window of 20 steps}
\label{RESL_20_smape}
\begin{tabular*}{\textwidth}{@{\extracolsep\fill}llrrrrrr}
\toprule%
& &\multicolumn{2}{@{}c@{}}{mean} & \multicolumn{2}{@{}c@{}}{90\%} & \multicolumn{2}{@{}c@{}}{divergence (\%)} \\\cmidrule{3-4}\cmidrule{5-6}\cmidrule{7-8}%
  &  & SETTR & OASIS & SETTR & OASIS & SETTR & OASIS \\
\midrule

1$\times \Delta t_{20}$ & $x$ & \textbf{2.63} & 2.66 & 3.92 & \textbf{2.24} & \textbf{7.14} & 55.36 \\
 & $y$ & \textbf{2.79} & 5.81 & \textbf{2.05} & 7.52 & \textbf{7.14} & 55.36 \\\midrule
2$\times \Delta t_{20}$ & $x$ & \textbf{6.21} & 7.35 & \textbf{11.32} & 24.77 & \textbf{7.14} & 50.00 \\
 & $y$ & \textbf{4.77} & 9.57 & \textbf{9.95} & 21.72 & \textbf{7.14} & 50.00 \\\midrule
3$\times \Delta t_{20}$ & $x$ & \textbf{9.82} & 11.83 & \textbf{23.89} & 25.61 & \textbf{7.14} & 44.64 \\
 & $y$ & \textbf{7.08} & 12.76 & \textbf{18.22} & 33.02 & \textbf{7.14} & 44.64 \\\midrule
4$\times \Delta t_{20}$ & $x$ & \textbf{9.88} & 12.51 & \textbf{16.45} & 35.09 & \textbf{3.57} & 53.57 \\
 & $y$ & \textbf{4.92} & 9.72 & \textbf{12.82} & 23.82 & \textbf{3.57} & 53.57 \\

\botrule
\end{tabular*}
\end{table}

\subsection{Detailed accuracy results for the comparison between Deep Set and Set Transformer}\label{appendix_2}

\begin{figure}[!htbp]
    \centering
    \includegraphics[width=1.\linewidth]{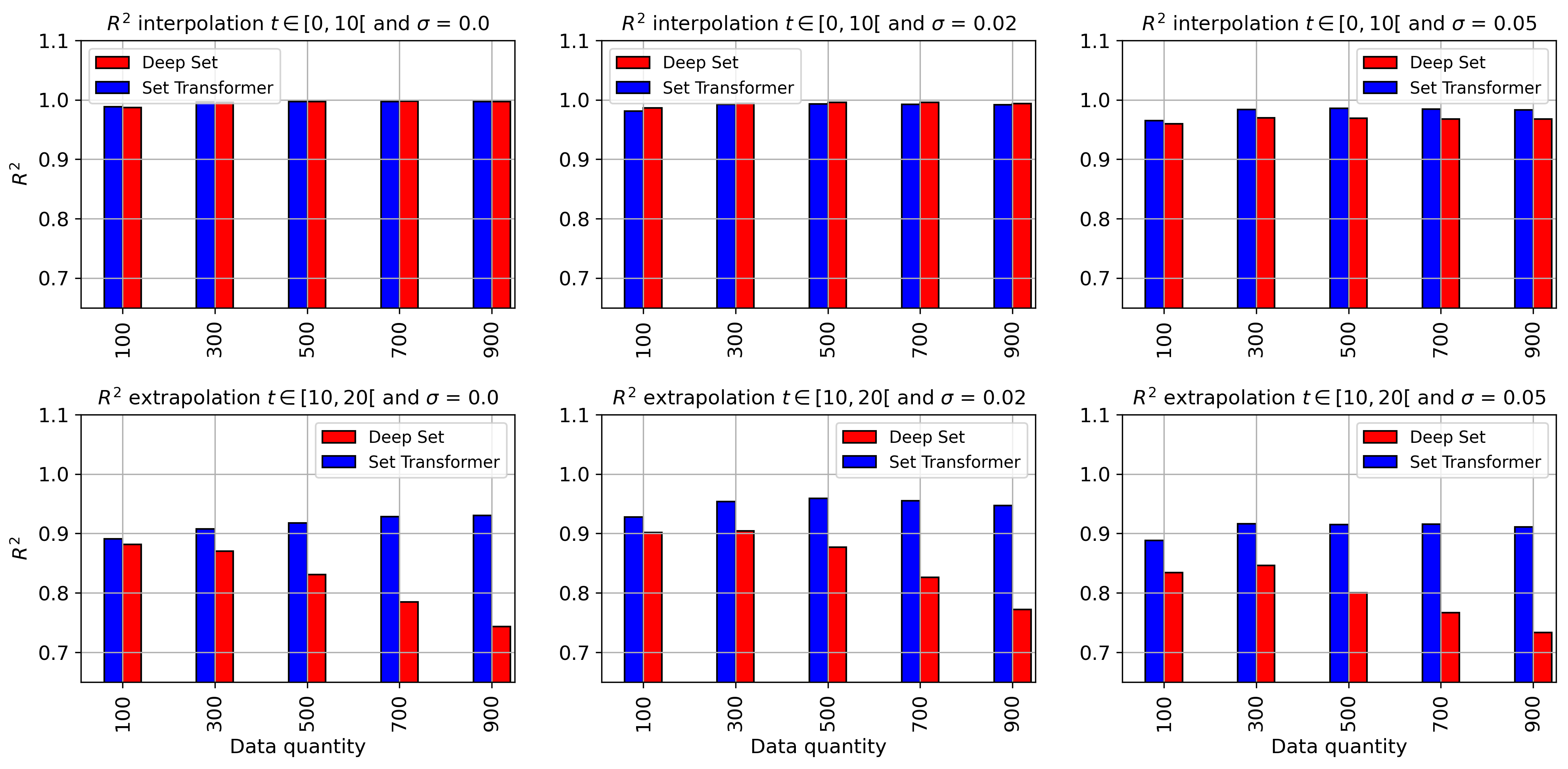}
    \caption{Accuracy comparison of $\sigma$ prediction between Deep Set and Set Transformer using the $R^2$ metric for the test dataset as a function of data quantity with different noise levels: 0 on the left, 0.02 in the center, and 0.05 on the right for interpolation (in the top) where $t\in[0,~10]$ and for extrapolation (in the bottom) where $t\in[10,~20]$.}
    \label{APP2SIGMA}
\end{figure}

\begin{figure}[!htbp]
    \centering
    \includegraphics[width=1.\linewidth]{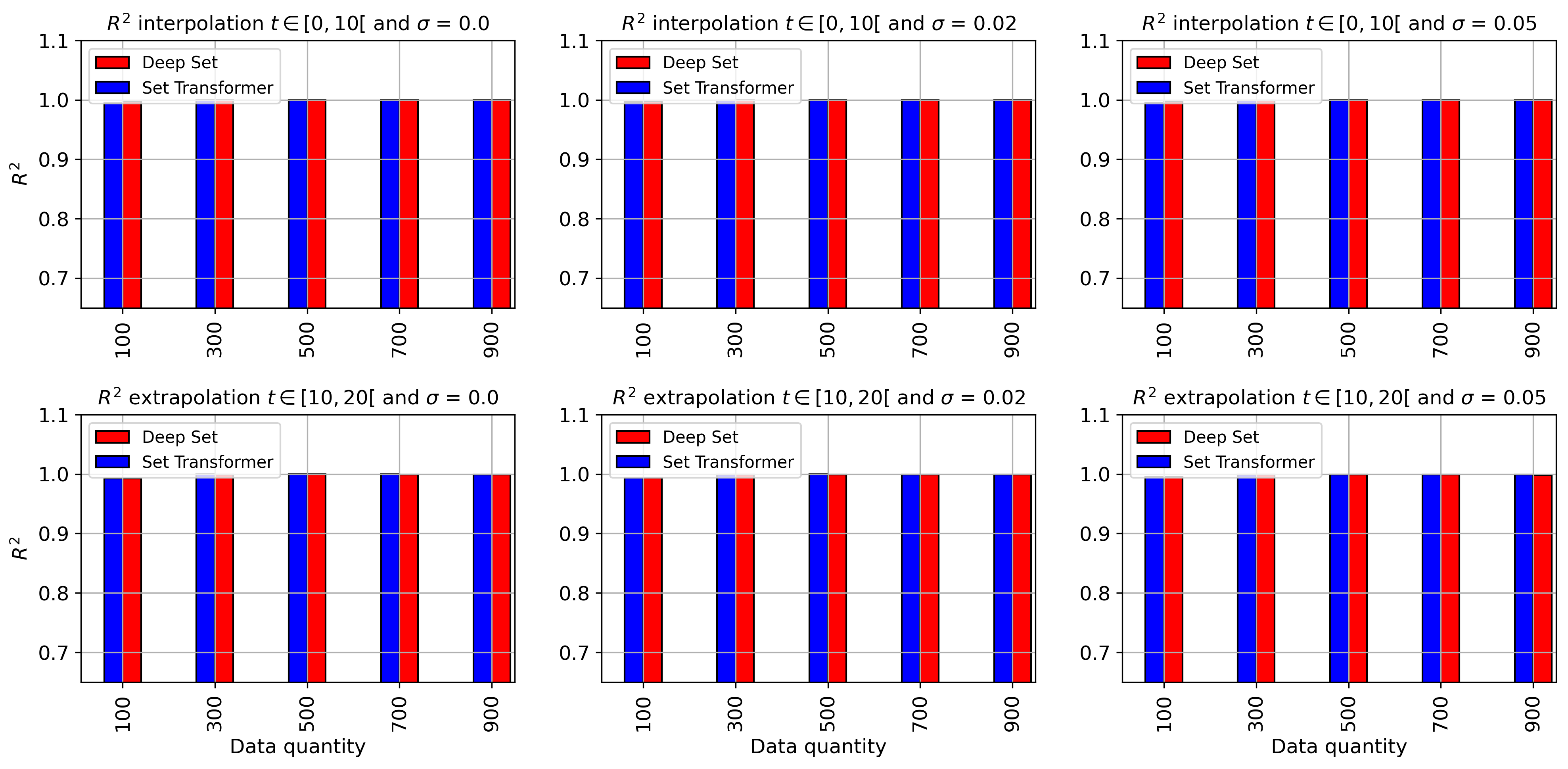}
    \caption{Accuracy comparison of $\beta$ prediction between Deep Set and Set Transformer using the $R^2$ metric for the test dataset as a function of data quantity with different noise levels: 0 on the left, 0.02 in the center, and 0.05 on the right for interpolation (in the top) where $t\in[0,~10]$ and for extrapolation (in the bottom) where $t\in[10,~20]$.}
    \label{APP2BETA}
\end{figure}

\begin{figure}[!htbp]
    \centering
    \includegraphics[width=1.\linewidth]{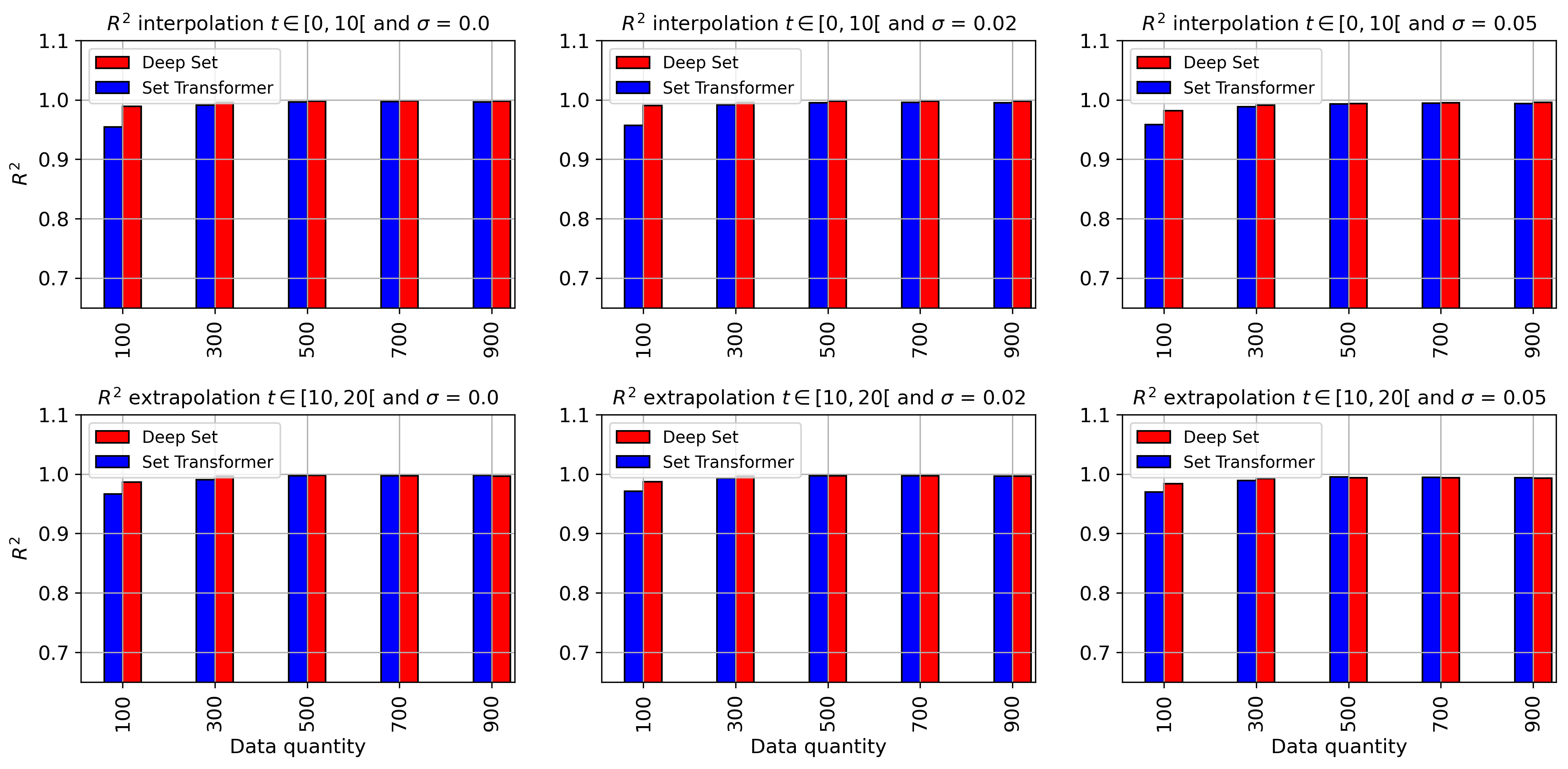}
    \caption{Accuracy comparison of $\rho$ prediction between Deep Set and Set Transformer using the $R^2$ metric for the test dataset as a function of data quantity with different noise levels: 0 on the left, 0.02 in the center, and 0.05 on the right for interpolation (in the top) where $t\in[0,~10]$ and for extrapolation (in the bottom) where $t\in[10,~20]$.}
    \label{APP2RHO}
\end{figure}

\textcolor{black}{
\subsection{Detailed model accuracy with respect to data quantity and noise levels}\label{appendix_4}
\begin{figure}[!htbp]
    \centering
    \includegraphics[width=0.8\linewidth]{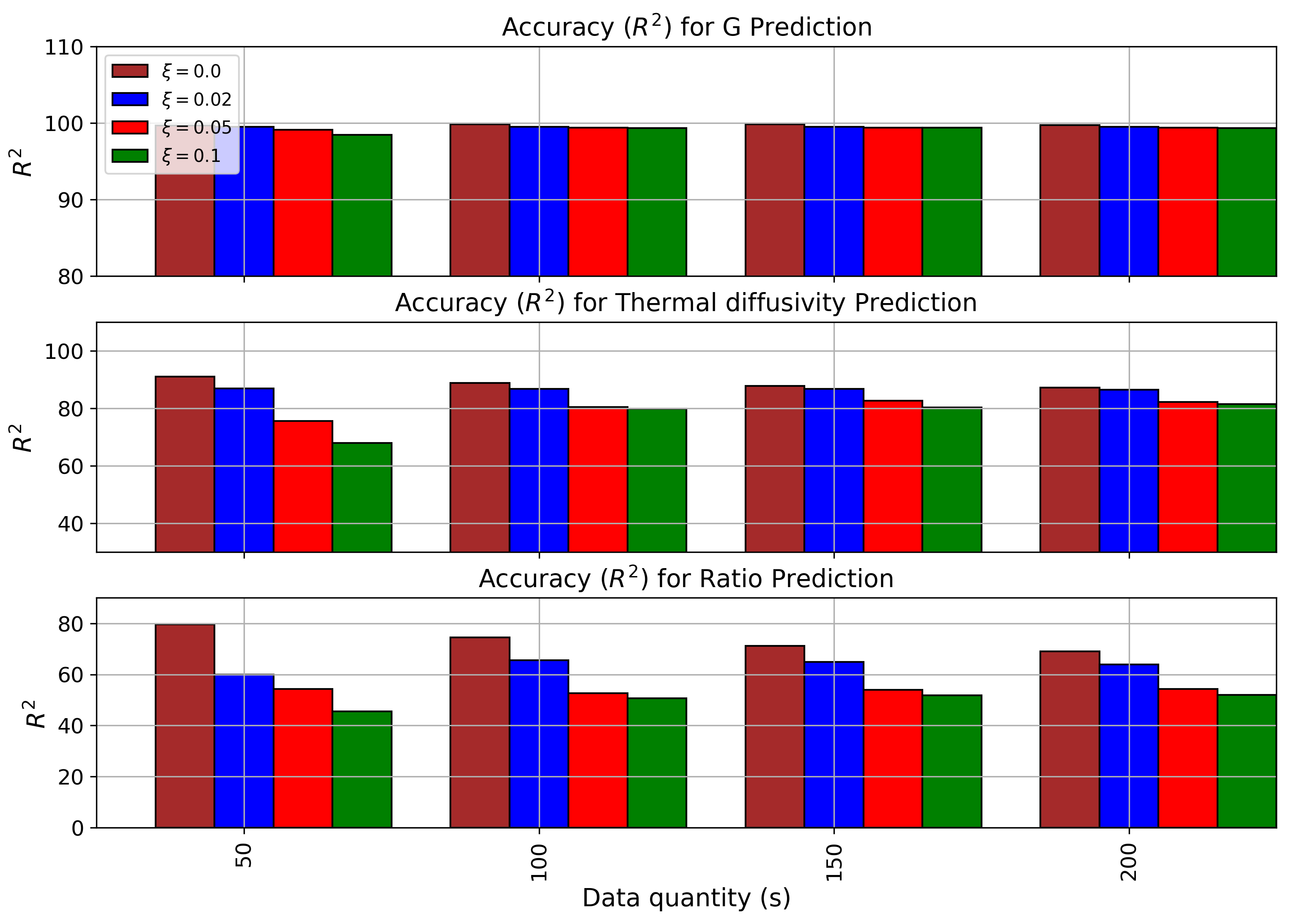}
    \caption{\textcolor{black}{Accuracy results using the $R^2$ metric and noise robustness with respect to data quantity for the Deep Set model for abnormalities characterization: $\mathrm{G}$, $\alpha_{ref}$, and $\mathrm{ratio}$, for four different noise levels ranging from $\xi = 0.0$ to $\xi = 0.1$.}}
    \label{APP3ALLRESL}
\end{figure}}

\newpage
\nocite{*}
    
\subsection{Instance selection for the comparison between Deep Set and Set Transformer}\label{appendix_3}
The two Set Encoding methods are trained on different architectures independently, considering three levels of noise and three parameters of the Lorenz system ($\sigma$, $\beta$, and $\rho$), with variable time windows. Simultaneously, they are validated on the validation dataset. A check-pointing process is used if the validation error (Tab.~\ref{tab:trainprocess2}) reaches a new minimum to save an instance of the architecture. The same process is conducted with different architectures for both methods, and the best instances are selected for each method, noise level, and each parameter of the Lorenz system. These instances are then used to compare the two methods. The selected architectures are presented in Tabs.~\ref{tab:optimal_dpst} and~\ref{tab:optimal_trfs}.\\

\textcolor{black}{\subsubsection{Deep Set}
The Deep Set model, as presented in section~\ref{SEC4}, can be summarized into three parts:
\begin{itemize}
\item The first part is the encoding part, which takes as input the state variables of the Lorenz system $[x(t),~y(t),~z(t)]$ and the time dimension $t$. It projects these state variables into a higher dimension using a simple MLP. The architecture of this MLP is defined in (Tab.~\ref{tab:optimal_dpst}), where we have 5 layers with 320 perceptrons each and a ReLU activation function.
\item The second part consists of choosing a pooling function over the time dimension. In this application, we found that applying the absolute value function followed by mean aggregation gives better results. The output of the aggregation function is a timeless feature vector.
\item The last part consists of decoding the feature vector into the dynamics parameters. In this application, we used an MLP with the same architecture as for encoding.
\end{itemize}}

\begin{table}[!htbp]
\centering\caption{Selected architecture for Deep Set method for all noise levels and all the parameters of the Lorenz system}
\label{tab:optimal_dpst}
\begin{tabular}{@{}llllc@{}}
\toprule
 & activation & Layers & Pooling                       & {\color[HTML]{0D0D0D} Trainable  parameter count} \\ \midrule
Encode & ReLU       & 320$\times5$  &                               &                                                   \\ 
Decode & ReLU       & 320$\times5$  & \multirow{-2}{*}{abs,   mean} & \multirow{-2}{*}{927043}                          \\ \botrule
\end{tabular}
\end{table}

\textcolor{black}{
\subsubsection{Set Transformer}
The Set Transformer architecture is more sophisticated compared to Deep Set, as it uses the attention mechanism, which means different encoding, pooling, and decoding approaches. Nevertheless, we can also summarize it as a three-part model:
\begin{itemize}
\item As for the Deep Set, this part consists of encoding the Lorenz system variables. For that, we used ISAB (section~\ref{SetTransformer}), which performs an attention mechanism encoding within the time and the state variables of the Lorenz system. These variables are projected into a higher dimension $d=45$, and then the attention mechanism is applied with respect to a trainable matrix $I$ called inducing points~\cite{SETTRANS}, with $m=128$ rows and $d=45$ columns. This means that ISAB will start by projecting the $n$ time snapshots into $m$ snapshots and then perform the attention mechanism with respect to the state variables of the Lorenz system. This method decreases the calculation time while providing similar results; more details are given in~\cite{SETTRANS}. For the attention mechanism, we used $h=40$ attention heads (Tab.~\ref{tab:optimal_dpst}). In this part, we apply a feedforward neural network to the output of the attention mechanism, adding more nonlinearity. For this application, the feedforward neural network is simply an MLP with 2 layers, 45 perceptrons each, and a ReLU activation function.
\item The second part consists of applying pooling over the encoder feature vector, which is based on PMA, a Multi-Head Attention block that performs aggregation using the multi-head attention mechanism. For this, we used a pooling seed vector $S \in \mathbb{R}^{k \times d}$, where $d=40$ and $k=1$.
\item Finally, we use an MLP with 2 layers, 40 perceptrons each, and a ReLU activation function, with a final layer of output dimension $d_{out}=3$ that corresponds to the 3 parameters of the Lorenz system.
\end{itemize}}

\begin{table}[!htbp]
\centering\caption{Selected architecture for Set Transformer method for all noise levels and all the parameters of the Lorenz system}
\label{tab:optimal_trfs}
\begin{tabular}{@{}lllc@{}}
\hline
 & Attention   blocks & rFF & {\color[HTML]{0D0D0D} Trainable   parameter count} \\ \hline
Encode & ISAB($d=45,   m=128, h=40$) & rFF(45, 2, ReLU) &  \\ 
Pooling & PAM($k=1, d=40, h=40$) & rFF(40, 2, ReLU) & \multirow{-2}{*}{1045733}  \\ \botrule
\end{tabular}
\end{table}

\textcolor{black}{
\subsection{Deep Set Architecture for the Third Application}\label{appendix_last}
The selected architecture is defined in Tab.~\ref{tab:optimal_dpst_app3}. As detailed in~\ref{appendix_3}, we selected the best model using the validation datasets and a checkpoint that saves instances when the validation loss reaches a new minimum. The selected architecture consists of three blocks. In the first block, we use an MLP with 5 layers, 256 perceptrons, and a GeLU activation function. For the second part, we use a sum aggregation function. The encoded vector is then decoded using three different MLPs with the same architecture as the encoder, each corresponding to one of the characteristics ($\mathrm{G}$, $\mathrm{ratio}$, and $\alpha_{ref}$).
\begin{table}[!htbp]
\centering
\caption{Selected architecture for the Deep Set method for all noise levels in the third application}
\label{tab:optimal_dpst_app3}
\begin{tabular}{@{}llllc@{}}
\toprule
 & Activation & Layers & Pooling & {\color[HTML]{0D0D0D} Trainable Parameter Count} \\ \midrule
Encode & GeLU & 256$\times$5 & & \\ 
Decode & GeLU & 3$\times$256$\times$5 & \multirow{-2}{*}{sum} & \multirow{-2}{*}{1262083} \\ \botrule
\end{tabular}
\end{table}}

\section*{Declarations}
\begin{itemize}
\item Funding : The Jules Verne Institute of Research and Technology (IRT Jules Verne) (PhD PERFORM Program).

\item Conflict of interest : We affirm that there are no known conflicts of interest that have influenced this work.

\item Ethics approval and consent to participate : We confirm that this research complies with authors ethical standards. All authors consent to participate in the submitted work.

\item Data availability
Not applicable.

\item Code availability
Not applicable.

\item Author contributions
Not applicable.

\end{itemize}

\bibliography{main}

\end{document}